\documentclass[journal]{IEEEtran}
\ifCLASSINFOpdf
\else
\fi
\usepackage{amsmath}
\usepackage{graphicx}
\usepackage{multirow}
\usepackage{subfigure}
\usepackage{caption}
\usepackage{rotating}
\usepackage{cite}
\usepackage{amssymb}
\usepackage{array}
\usepackage{bm}
\usepackage{color}
\usepackage{marvosym}
\usepackage{pifont}

\definecolor{green}{RGB}{0,139,69}
\usepackage[colorlinks,
            linkcolor=blue,
            anchorcolor=blue,
            citecolor=blue]{hyperref}

\usepackage{hyperref}
\makeatletter
\def\UrlAlphabet{%
      \do\a\do\b\do\c\do\d\do\e\do\f\do\g\do\h\do\i\do\j%
      \do\k\do\l\do\m\do\n\do\o\do\p\do\q\do\r\do\s\do\t%
      \do\u\do\v\do\w\do\x\do\y\do\z\do\A\do\B\do\C\do\D%
      \do\E\do\F\do\G\do\H\do\I\do\J\do\K\do\L\do\M\do\N%
      \do\O\do\P\do\Q\do\R\do\S\do\T\do\U\do\V\do\W\do\X%
      \do\Y\do\Z}
\def\UrlDigits{\do\1\do\2\do\3\do\4\do\5\do\6\do\7\do\8\do\9\do\0}
\g@addto@macro{\UrlBreaks}{\UrlOrds}
\g@addto@macro{\UrlBreaks}{\UrlAlphabet}
\g@addto@macro{\UrlBreaks}{\UrlDigits}
\makeatother

\newcommand{\tabincell}[2]{\begin{tabular}{@{}#1@{}}#2\end{tabular}}

\newcommand{\ie}{\textit{i}.\textit{e}.}
\newcommand{\eg}{\textit{e}.\textit{g}.}

\newcommand{\figref}[1]{Fig.~\ref{#1}}

\newcommand{\secref}[1]{Section~\ref{#1}}

\graphicspath{{./Imgs/}}
\DeclareGraphicsExtensions{.jpg,.pdf,.png}
\begin{document}
\title{Light Field Salient Object Detection:\\
A Review and Benchmark}

\author{Keren Fu,
        Yao~Jiang,
        Ge-Peng~Ji,
        Tao~Zhou*,
        Qijun Zhao,
        and~Deng-Ping Fan% <-this % stops a space
\thanks{K. Fu, Y. Jiang, and Q. Zhao are with the College of Computer Science, Sichuan University, China. (Email: fkrsuper@scu.edu.cn, yaojiangyj@foxmail.com, qjzhao@scu.edu.cn).}% <-this % stops a space
\thanks{G.-P. Ji is with the School of Computer Science, Wuhan University, China. (Email: gepengai.ji@gmail.com)}
\thanks{T. Zhou is with PCA Lab, School of Computer Science and Engineering, Nanjing University of Science and Technology, China. (Email: taozhou.ai@gmail.com).}
\thanks{D.-P. Fan is with CS, Nankai University, Tianjin 300350, China. (Email: dengpfan@gmail.com).}

% <-this % stops a space
\thanks{Corresponding author: Tao~Zhou.}}


\markboth{Journal of LaTeX Class Files,~Vol.~14, No.~8, August~2015}%
{Shell \MakeLowercase{\textit{et al.}}: Bare Demo of IEEEtran.cls for IEEE Journals}



\maketitle

\begin{abstract}
Salient object detection (SOD) is a long-standing research topic in computer vision and has drawn an increasing amount of research interest in the past decade.
Since light fields record comprehensive information of natural scenes that benefit SOD in a number of ways, using the light field inputs to improve saliency detection over conventional RGB inputs is an emerging trend. This paper provides the first comprehensive review and benchmark for light field SOD, which has long been lacking in the saliency community.
Firstly, we introduce preliminary knowledge on light fields, including theory and data forms, and then review existing studies on light field SOD, covering ten traditional models, seven deep learning-based models, one comparative study, and one brief review. Existing datasets for light field SOD are also summarized with detailed information and statistical analyses.
Secondly, we benchmark nine representative light field SOD models together with several cutting-edge RGB-D SOD models on four widely used light field datasets, from which insightful discussions and analyses, including a comparison between light field SOD and RGB-D SOD models, are achieved. Besides, due to the inconsistency of datasets in their current forms, we further generate complete data and supplement focal stacks, depth maps and multi-view images for the inconsistent datasets, making them consistent and unified. Our supplemental data makes a universal benchmark possible.
Lastly, because light field SOD is quite a special problem attributed to its diverse data representations and high dependency on acquisition hardware, making it differ greatly from other saliency detection tasks, we provide nine hints into the challenges and future directions, and outline several open issues. We hope our review and benchmarking could help advance research in this field.
All the materials including collected models, datasets, benchmarking results, and supplemented light field datasets will be publicly available on our project site \url{https://github.com/kerenfu/LFSOD-Survey}.
\end{abstract}

\begin{IEEEkeywords}
Light field, salient object detection, deep learning, benchmarking.
\end{IEEEkeywords}

% \fi
\IEEEpeerreviewmaketitle

\section{Introduction}
In Google I/O 2021 held in May, Google just introduced its new technology called Project Starline\footnote{\url{https://blog.google/technology/research/project-starline/}}, which combines specialized hardware and computer vision technology to create a ``magic window'' that can reconnect two remote persons, making them feel as if they are physically sitting in front of each other in the conversation. Such immersive technology, is benefited from \emph{light field} display, and does not require additional glasses or headsets. The three crucial techniques involved therein, include 3D imaging, real-time data compression, and light field-based 3D display, which are very challenging but have had certain breakthroughs according to Google. In this regard, salient object detection (SOD) from the light field may be another effective way \cite{Wang2018SalienceGDSGDC} to benefit these three stages.

Salient object detection (SOD)~\cite{Cheng2011GlobalCB,Borji2015SalientOD,Borji2019SalientOD} is a fundamental task in computer vision, aiming at detecting and segmenting conspicuous regions or objects in a scene, whereas light field SOD \cite{Li2014SaliencyDOLFS,li2017saliency} studies the problem of how to realize SOD using light field data. Numerous applications of SOD cover, \eg, object detection and recognition~\cite{Ren2014RegionBasedSD,Zhang2016BridgingSD,Rutishauser2004IsBA,Moosmann2006LearningSM,cheng2019bing}, semantic segmentation~\cite{Wei2017ObjectRM,Wei2017STCAS,Wang2018WeaklySupervisedSS}, unsupervised video object segmentation~\cite{Wang2018SaliencyAwareVO,Song2018PyramidDD}, multimedia compression~\cite{Itti2004AutomaticFF,Ma2005AGF,Ma2002AUA,Ouerhani2001AdaptiveCI}, non-photorealist rendering~\cite{Han2013FastSM}, re-targeting~\cite{Sun2011ScaleAO}, and also human-robot interaction~\cite{Sugano2010CalibrationfreeGS,Borji2014DefendingYE}. Generally, the abundant cues and information within the light field help algorithms better identify target objects and advance SOD performance comparing to the conventional SOD task that processes single-colour images~\cite{Fu2019DeepsideAG,Wang2016CorrespondenceDS,Zhang2017AmuletAM,Feng2019AttentiveFN,Zhang2017LearningUC}.

\begin{figure}[t]
\centering
\includegraphics[width=0.48\textwidth]{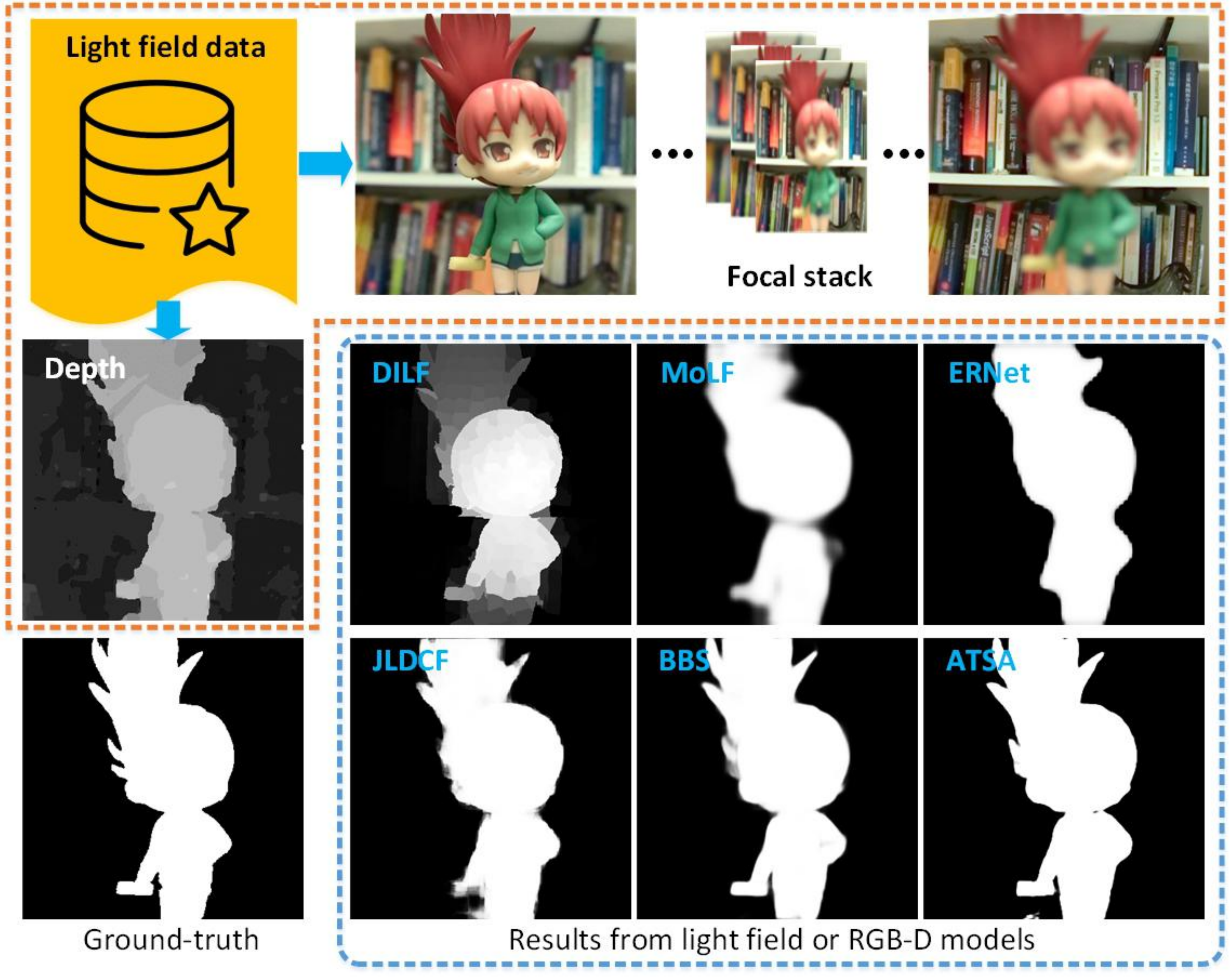}
\caption{Salient object detection on a sample scenario using three light field SOD models (\ie, DILF~\cite{Zhang2015SaliencyDILF}, MoLF~\cite{Zhang2019MemoryorientedDFMoLF} and ERNet~\cite{Piao2020ExploitARERNet}), and three state-of-the-art RGB-D SOD models (\emph{i.e.}, JLDCF~\cite{Fu2020JLDCFJL,Fu2021siamese}, BBS~\cite{Fan2020BBSNetRS}, and ATSA~\cite{Zhang2020ASTA}).}
\label{light field and RGB-D examples}
\end{figure}

%In recent years, deep learning-based SOD has shown great potential, and attracted increasing research interest. However, SOD based on a single modality (\ie, detection on a single RGB input image) still encounters several challenges in complex scenarios with, for instance, heavy background clutter~\cite{fan2018salient}, or high similarity between foreground and background. To solve these challenges, incorporating additional supplementary knowledge, such as scene depth~\cite{Fan2020BBSNetRS,Fu2020JLDCFJL,Zhang2020SelectSA,Zhang2020UCNetUI,Fan2020RethinkingRS,LiuLearningSS,Li2020RGBDSO,Pang2020HierarchicalDF,Zhang2020ASTA,Ge2021WGINetAW} or motion~\cite{Li2019MotionGA}, has been shown to boost SOD performance. In this regard, using the light field~\cite{Li2014SaliencyDOLFS,Li2015WSC,Zhang2015SaliencyDILF,Sheng2016RelativeLFRL,Wang2017BIF,li2017saliency,zhang2017MA,Wang2018AccurateSSDDF,Wang2018SalienceGDSGDC,Wang2020RegionbasedDRDFD,Piao2020SaliencyDVDCA,Wang2019DeepLFDLLF,Piao2019DeepLSDLSD,Zhang2019MemoryorientedDFMoLF,Piao2020ExploitARERNet,Zhang2020LFNetLFNet,Zhang2020LightFSLFDCN,zhang2020multiMTCNet,Zhang2015LightFSCS} is another emerging trend.

Light field SOD explores how to detect salient objects using light field data as input.
In the 3D space, a light field~\cite{Gershun1939TheLF} captures all the light rays at every spatial location and in every direction. As a result, it can be viewed as an array of images captured by a grid of cameras.
Compared with the RGB images captured by a regular camera or depth maps acquired by a depth sensor, the light field data acquired by a plenoptic camera records more comprehensive and complete information of natural scenes, covering, for example, depth information~\cite{Jeon2015AccurateDM,Tao2013DepthFC,Tao2015DepthFS,Wang2015OcclusionAwareDE,Ng2005LightFP,Piao2021DynamicFN,Piao2021LearningMI}, focusness cues~\cite{Ng2005LightFP,Li2014SaliencyDOLFS} as well as angular changes~\cite{Ng2005LightFP,Piao2019DeepLSDLSD}.
Therefore, light field data can benefit SOD in a number of ways. Firstly, light fields can be refocused after being acquired~\cite{Ng2005LightFP}. This enables a stack of images focusing on different depths to be produced, providing focusness cues that have been demonstrated useful for SOD~\cite{Jiang2013SalientRD}.
Secondly, a light field can provide images of a scene from an array of viewpoints~\cite{Buehler2001UnstructuredLR}. Such images have abundant spatial parallax and geometric information. Lastly, the depth information of a scene is embedded in light field data and can be estimated from a focal stack or multi-view images by different means, as described in~\cite{Jeon2015AccurateDM,Tao2013DepthFC,Tao2015DepthFS,Wang2015OcclusionAwareDE}. In this sense, RGB-D data could be deemed as a special degenerated case of light field data. Fig. \ref{light field and RGB-D examples} shows example results obtained using light field SOD methods on light field data (a focal stack), as well as RGB-D SOD models on depth data, respectively.

Although light field data brings great benefits to SOD and this task first emerged in 2014~\cite{Li2014SaliencyDOLFS}, it still remains somewhat under-explored. Specifically, compared with RGB SOD or RGB-D SOD, there are fewer studies on light field SOD.
%there are only \fkr{18 individual} studies regarding SOD on light fields.
Despite this sparsity of literature, existing models vary in technical frameworks as well as light field datasets used. However, to the best of our knowledge, there is no comprehensive review or benchmark for light field SOD. Although a comparative study was conducted by Zhang~\emph{et al.}~\cite{Zhang2015LightFSCS} in 2015, they only compared the classic light field SOD model proposed by Li~\emph{et al.}~\cite{Li2014SaliencyDOLFS} with a set of 2D saliency models to demonstrate the effectiveness of incorporating light field knowledge. Besides, the evaluation was conducted on the LFSD dataset, which only contains 100 light field images.
Recently, Zhou~\emph{et al.}~\cite{Zhou2020RGBDSO} briefly summarized existing light field SOD models and related datasets. However, their work was mainly focused on RGB-D based SOD, and only a small portion of content and space was dedicated to reviewing light field SOD, leading to an insufficient review of model details and related datasets. Besides, they did not benchmark light field SOD models or provide any performance evaluation. Thus, we believe that the lack of a complete review of the existing models and datasets may hinder further research in this field.

To this end, in this paper, we conduct \emph{the first comprehensive review and benchmark for light field SOD}. We review previous studies on light field SOD, including ten traditional models~\cite{Li2014SaliencyDOLFS,Li2015WSC,Zhang2015SaliencyDILF,Sheng2016RelativeLFRL,Wang2017BIF,zhang2017MA,Wang2018SalienceGDSGDC,Wang2020RegionbasedDRDFD,Piao2020SaliencyDVDCA,Wang2018AccurateSSDDF}, seven deep learning-based models~\cite{Wang2019DeepLFDLLF,Piao2019DeepLSDLSD,Zhang2019MemoryorientedDFMoLF,Piao2020ExploitARERNet,Zhang2020LFNetLFNet,Zhang2020LightFSLFDCN,zhang2020multiMTCNet}, one comparative study~\cite{Zhang2015LightFSCS}, and one brief review~\cite{Zhou2020RGBDSO}. In addition, we also review existing light field SOD datasets~\cite{Li2014SaliencyDOLFS,zhang2017MA,Wang2019DeepLFDLLF,Piao2019DeepLSDLSD,Zhang2020LightFSLFDCN}, and provide statistical analyses for them, covering object size, the distance between the object and image center, focal slice number, and a number of objects. Due to the inconsistency of datasets (for example, some do not provide focal stacks, while others lack depth maps or multi-view images), we further generate and complete data, including focal stacks, depth maps and multi-view images for several datasets, therefore making them consistent and unified.
Besides, we benchmark nine light field SOD models~\cite{Li2014SaliencyDOLFS,Li2015WSC,Zhang2015SaliencyDILF,Wang2020RegionbasedDRDFD,Piao2019DeepLSDLSD,Piao2020ExploitARERNet,Zhang2019MemoryorientedDFMoLF,Zhang2020LFNetLFNet,Zhang2020LightFSLFDCN} whose results/codes are available, together with several cutting-edge RGB-D SOD models~\cite{Fan2020BBSNetRS,Fu2020JLDCFJL,Zhang2020SelectSA,Zhang2020UCNetUI,Fan2020RethinkingRS,LiuLearningSS,Li2020RGBDSO,Pang2020HierarchicalDF,Zhang2020ASTA}, discussing the connection between the two and providing insight into the challenges and future directions. All the materials involved in this paper, including collected models, benchmark datasets, and results, supplemental light field data, and source code links, will be made publicly available on \url{https://github.com/kerenfu/LFSOD-Survey}.
The main contributions of this paper are four-fold: %are summarized as follows:
\begin{itemize}

    \item We provide the first systematic review on light field SOD, including models and datasets. Such a survey has long been lacking in the saliency community and is helpful for encouraging future research in this area.

    \item We conduct analyses on the properties of different datasets. As some datasets lack certain forms of data, \emph{e.g.}, focal stacks, or multi-view images, we generate more data from existing datasets as a supplement, making them complete and unified. This will also facilitate future research in this area.

    \item We further benchmark nine light field SOD models together with several cutting-edge RGB-D SOD models on the datasets with our supplemental data, and provide insightful discussions.

    \item We investigate several challenges for light field SOD and discuss its relation to other topics, shedding light on the challenges and directions for future work.
\end{itemize}

The remainder of the paper is organized as follows. We review light field preliminaries, existing models and datasets for light field SOD, and provide related discussions and analyses in~\secref{sec:priors_of_LFSOD}. In~\secref{sec:model_eval_benchmarks}, we describe the  evaluation metrics used as well as benchmark results. We then discuss future research directions and outline several open issues of this field in~\secref{challenge and open direction}. Finally, we draw conclusion in \secref{sec:conclusions}.

\section{Preliminaries, Models and Datasets} \label{sec:priors_of_LFSOD}

In this section, we first briefly introduce the theory of light field, the data forms of it, and how it has been used for SOD. We then review the previous works on light field SOD, roughly categorizing them into traditional models and deep learning-based models. Finally, we summarize datasets explored for light field SOD and review their detailed information.

\begin{figure}
\centering
\includegraphics[width=0.48\textwidth]{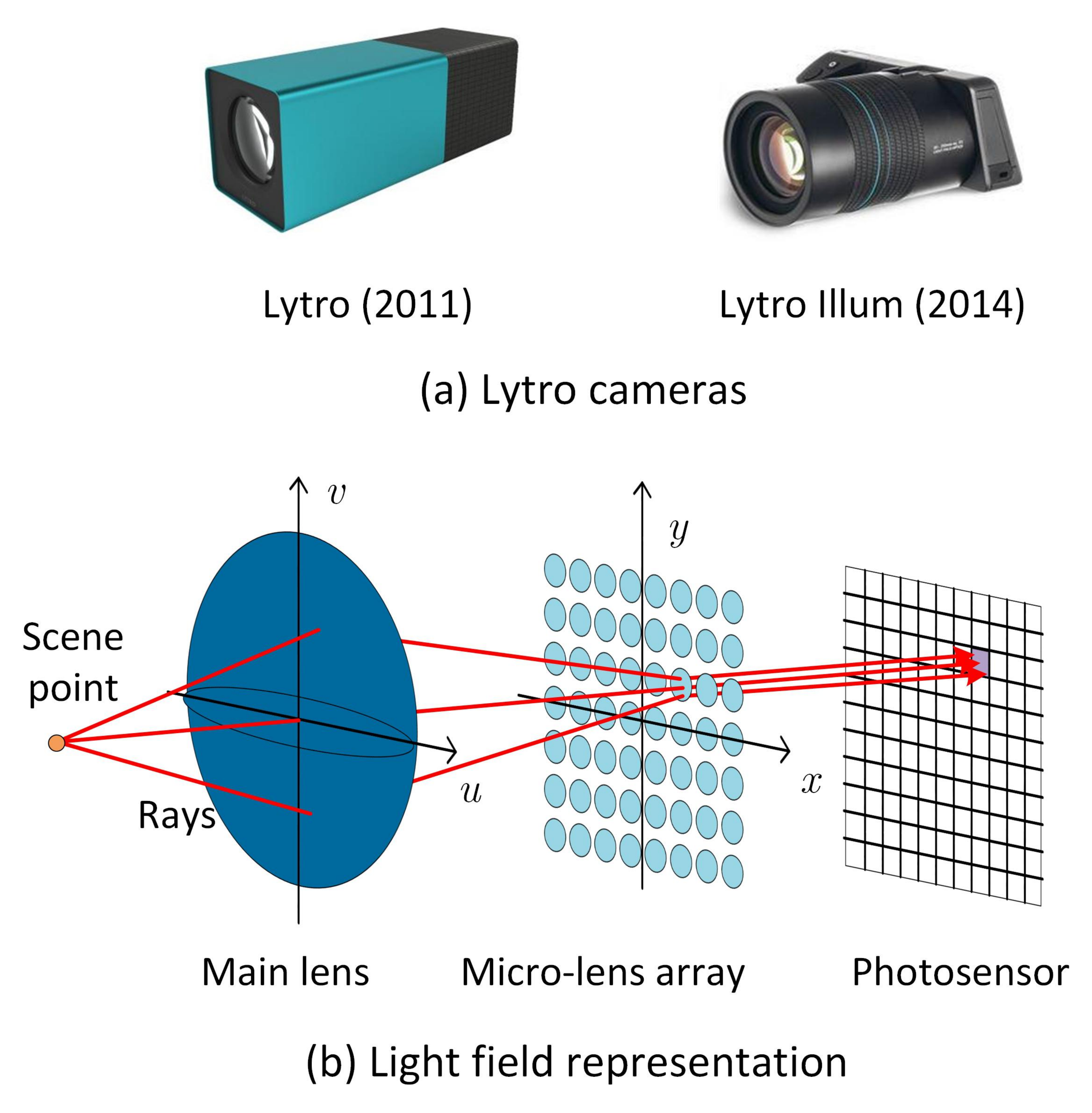}
\caption{Lytro cameras (a) and representation of light field (b).}
\label{Lytro camera and representation of light field}
\end{figure}

\subsection{Light Field}

\subsubsection{Light Field and Light Field Cameras}

A light field~\cite{Gershun1939TheLF} consists of all the light rays flowing through every point and in every direction of the 3D space. In 1991, Adelson and Bergen~\cite{Landy1991ThePF} proposed a plenoptic function ($P$), which uses $P(\theta,\phi,\lambda,t,x,y,z)$ to describe the wavelength $\lambda$ and time $t$ in any direction $(\theta,\phi)$ at any point $(x,y,z)$, in order to represent the light field information. In an imaging system, the wavelength and time can be represented by RGB channels and different frames, and light usually propagates along a specific path. As a result, Levoy and Hanrahan~\cite{Levoy1996LightFR} proposed the two-plane parameterization of the plenoptic function to represent the light field in an imaging system.
The two-plane parameterization ($L$) of the plenoptic function, illustrated in Fig. \ref{Lytro camera and representation of light field} (b), can be formulated as $L(u,v,x,y)$. In this scheme, each ray in the light field is determined by two parallel planes to represent spatial $(x,y)$ and angular $(u,v)$ information. Based on this theory, devices that can capture light fields were invented, namely the Lytro cameras shown in Fig. \ref{Lytro camera and representation of light field} (a). This kind of camera contains the main lens and a micro-lens array placed before the photosensor, where the former serves as the ``$uv$'' plane, which records the angular information of rays, and the latter serves as the ``$xy$'' plane, which records the spatial information. Fig. \ref{Lytro camera and representation of light field} (b) shows the graphical representation of the two-plane parameterization for the light field. Due to the above four-dimensional parameterization, data from such a light field is often called 4D light field data~\cite{Li2014SaliencyDOLFS,Li2015WSC,Zhang2015SaliencyDILF,Sheng2016RelativeLFRL,Wang2017BIF,li2017saliency,zhang2017MA,Wang2018SalienceGDSGDC,Wang2020RegionbasedDRDFD,Piao2020SaliencyDVDCA,Wang2019DeepLFDLLF,Piao2019DeepLSDLSD,Zhang2019MemoryorientedDFMoLF,Piao2020ExploitARERNet,Zhang2020LFNetLFNet,Zhang2020LightFSLFDCN,zhang2020multiMTCNet,Zhang2015LightFSCS,Wang2018AccurateSSDDF}.

\subsubsection{Forms of Light Field Data}
\label{Forms of Light Field Data}

\begin{figure}
\centering
\includegraphics[width=0.48\textwidth]{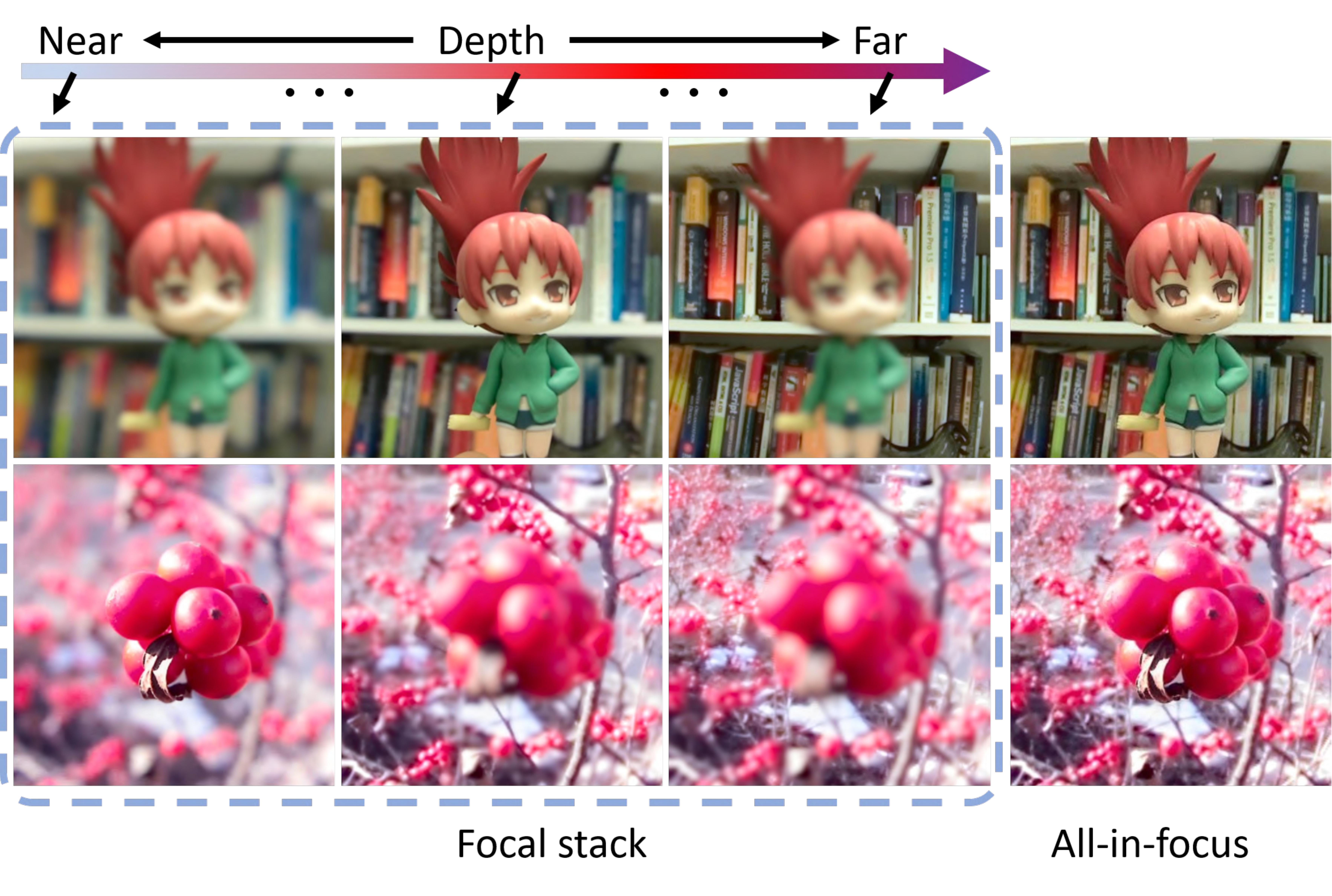}\vspace{-0.3cm}
\caption{Focal stacks and all-in-focus images.}
\label{focal stack and all focus images}
\end{figure}

Up to now, all public light field datasets for SOD have been captured by Lytro cameras, the raw data of which are LFP/LFR files (the former are obtained from Lytro whereas the latter are from Lytro Illum). All images in the current light field datasets were generated by processing LFP/LFR files using Lytro Desktop software\footnote{\url{http://lightfield-forum.com/lytro/lytro-archive/}} or LFToolbox\footnote{\url{http://code.behnam.es/python-lfp-reader/} and also \url{https://ww2.mathworks.cn/matlabcentral/fileexchange/75250-light-field-toolbox}}. Since the raw data can hardly be utilized, the data forms of light fields leveraged by existing SOD models are diverse, including focal stacks plus all-in-focus images~\cite{Li2014SaliencyDOLFS,Li2015WSC,Zhang2015SaliencyDILF,Wang2017BIF,zhang2017MA,Wang2018SalienceGDSGDC,Wang2020RegionbasedDRDFD,Piao2020SaliencyDVDCA,Piao2019DeepLSDLSD,Zhang2019MemoryorientedDFMoLF,Piao2020ExploitARERNet,Zhang2020LFNetLFNet}, multi-view images plus center-view images~\cite{Piao2019DeepLSDLSD,zhang2017MA,zhang2020multiMTCNet}, and micro-lens image arrays~\cite{Sheng2016RelativeLFRL,Zhang2020LightFSLFDCN}. As mentioned before, depth images can also be synthesized from light field data~\cite{Jeon2015AccurateDM,Tao2013DepthFC,Tao2015DepthFS,Wang2015OcclusionAwareDE}, and therefore they can form RGB-D data sources for RGB-D SOD models (Fig. \ref{light field and RGB-D examples}). Focal stacks and all-in-focus images are shown in Fig. \ref{focal stack and all focus images}, whereas multi-view images, center-view images and depth images are shown in Fig. \ref{multi-view images, depth and ground truth}.

\begin{figure}
\centering
\includegraphics[width=0.48\textwidth]{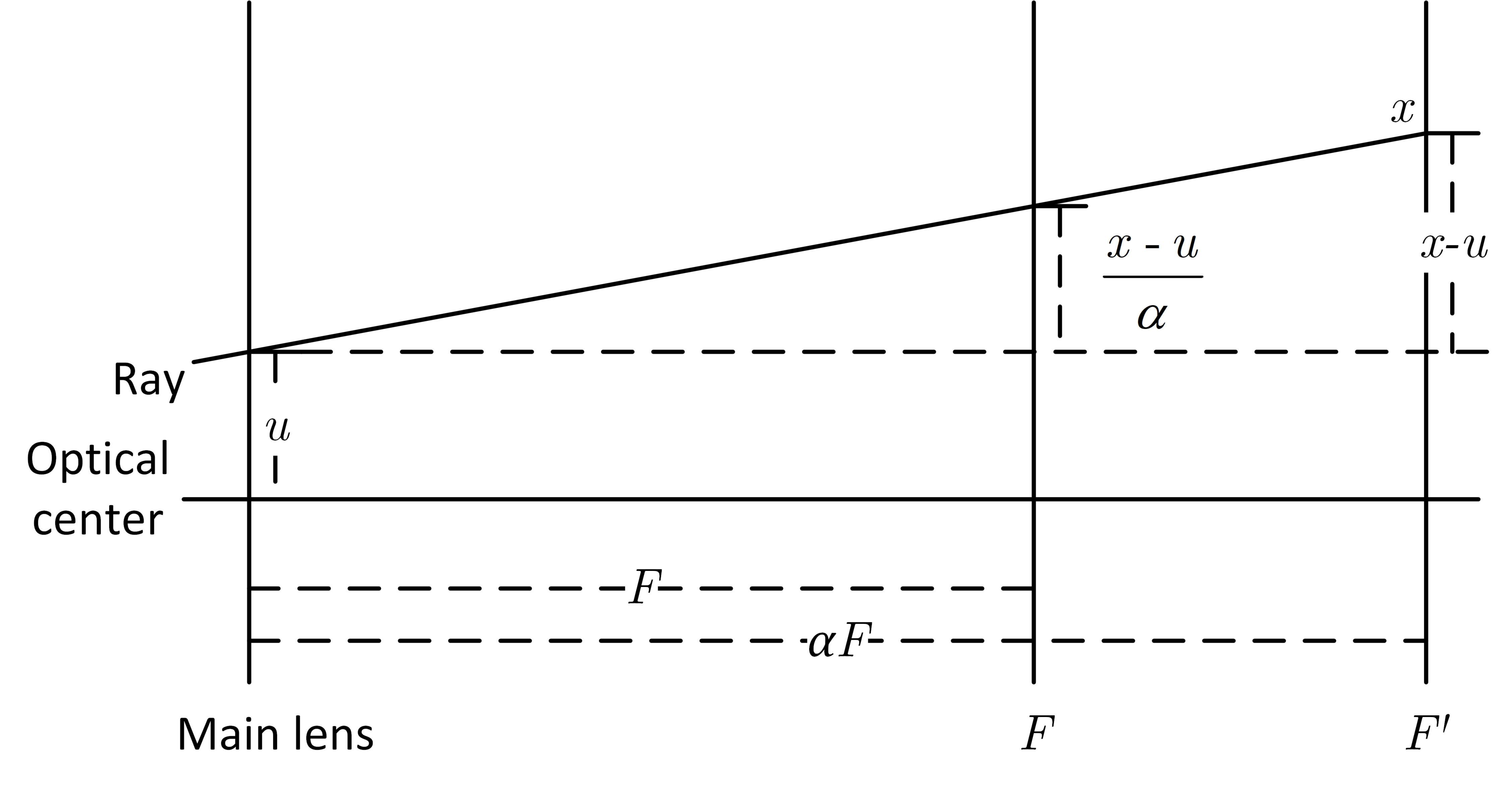}
\caption{Illustration of the refocus principle.}
\label{refocus principle}
\end{figure}

Specifically, a focal stack (left three columns in~\figref{focal stack and all focus images}) contains a series of images focusing on different depths. Such images are generated by processing the raw light field data using digital refocusing techniques~\cite{Ng2005LightFP}. The refocusing principle is demonstrated in Fig. \ref{refocus principle}, which only shows the case of $u$ and $x$ dimensions.
Suppose a light ray enters the main lens at location $u$. If the imaging plane's position $F$ ($F$ denotes the focal distance of the main lens) is changed to $F'$, where $F' = \alpha F$, a refocused image can be computed as follows. First, given the 4D light field $L_F$, the new light field $ L_\alpha$ regarding the new imaging plane at $F'$ can be derived as
\begin{align}
    L_\alpha(u,v,x,y) = L_F(u,v,u+\frac{x-u}{\alpha},v+\frac{y-v}{\alpha}).
\end{align}
Next, after obtaining the new light field $L_\alpha(u,v,x,y)$, a refocused image on the imaging plane can be synthesized as
\begin{align}
    I_\alpha(x,y) = \iint L_\alpha(u,v,x,y)du,dv.
\end{align}
One can see that by changing the parameter $\alpha$, a series of refocused images can be generated, composing a focal stack. After the focal stack is obtained, an all-in-focus image can be produced by photo-montage~\cite{Agarwala2004InteractiveDP}.
For example, an all-in-focus image can be generated by putting all the clear pixels together, where the clarity of pixels can be estimated by the associated gradients. Besides, it can also be acquired by computing a weighted average of all the focus slices. More details of the algorithm can be found in~\cite{Kuthirummal2011FlexibleDO}.

In addition to focal stacks, multi-view images (Fig. \ref{multi-view images, depth and ground truth}) can also be derived from light field data. As mentioned before, in the 4D light field representation $L_F(u,v,x,y)$, $(u,v)$ encode the angular information of incoming rays. Thus, an image from a certain viewpoint can be generated by sampling at a specific angular direction $(u^*,v^*)$, and the image can be represented by $L_F(u^*,v^*,x,y)$. By varying $(u^*,v^*)$, multi-view images can be synthesized. Especially, when the angular direction $(u^*,v^*)$ is equal to that of the central view, namely $(u_0,v_0)$, the center-view image is achieved. On the other hand, micro-lens images can be generated by sampling the $(x,y)$ dimensions. Providing a micro-lens location $(x^*, y^*)$ leads to a micro-lens image $L_F(u,v,x^*,y^*)$, which captures multiple perspectives of a scene point. Note that, by varying $(x^*, y^*)$, different micro-lens images can be obtained, which together compose a micro-lens image array representing complete light field information. A visualization of micro-lens and multi-view images can be found in a recent work~\cite{Zhang2020LightFSLFDCN}.

Moreover, depth maps containing scene depth information can also be estimated from a light field. As mentioned, depth information is embedded in the focusness and angular cues, so a depth map can be generated by combining both defocusness and correspondence (angular) cues. For more details on depth estimation from light fields please refer to~\cite{Jeon2015AccurateDM,Tao2013DepthFC,Tao2015DepthFS,Wang2015OcclusionAwareDE,Piao2021DynamicFN,Piao2021LearningMI}.

\subsection{Light Field SOD Models and Reviews}\label{Light Field SOD Models and Reviews}

In this section, we review and discuss existing models proposed for light field SOD, including ten traditional models that employ hand-crafted features, and seven deep learning-based models. Also, one comparative study and one brief review are revisited. Details of all these works are summarized in Table \ref{model table}.

\begin{table*}[!ht]
\centering
\caption{Overview of light field SOD models and review works. For datasets, please see Table \ref{dataset table}. We use the following abbreviations: FS=focal stacks, DE=depth maps, MV=multi-view images, ML=micro-lens images, OP=open-source. FS, DE, MV and ML indicate the data forms input to the models. Emerged datasets are highlighted in \textbf{bold} in the ``Main components''.}
\footnotesize
\renewcommand{\arraystretch}{0.9}
\renewcommand{\tabcolsep}{0.8mm}
\resizebox{\textwidth}{8.5cm}{
\begin{tabular}{|c|l||cc||c|c|p{5cm}|cccc|c|}
\hline
 & Models                                      & Pub.   & Year & Training dataset(s)    & Testing dataset(s)           & Main components & FS       &  DE      &  MV             &  ML      &  OP        \\
\hline
\hline
& LFS~\cite{Li2014SaliencyDOLFS}               & CVPR   & 2014 &  -               & LFSD                   & Focusness measure, location priors, contrast cues, background prior, \textbf{new dataset (LFSD)}                &\ding{51}&          &                 &          &\ding{51}                                         \\
\cline{2-12}
& WSC~\cite{Li2015WSC}                         & CVPR   & 2015 &  -               & LFSD                   &  Weighted sparse coding,  saliency/non-saliency dictionary construction      &\ding{51}&    \ding{51}      &                 &          &\ding{51}                                         \\
\cline{2-12}
\multirow{12}{*}{\begin{sideways} \textbf{Traditional models}\end{sideways}}&DILF~\cite{Zhang2015SaliencyDILF}            & IJCAI  & 2015 &  -               & LFSD                   &   Depth-induced/Color contrast, background priors by focusness              &\ding{51}&\ding{51}&                 &          &\ding{51}                                       \\
\cline{2-12}
&RL~\cite{Sheng2016RelativeLFRL}              & ICASSP & 2016 &  -               & LFSD                   &  Relative locations, guided filter, micro-lens images\qquad\qquad\qquad\qquad&          &          &                 &\ding{51}&                                               \\
\cline{2-12}
&BIF~\cite{Wang2017BIF}                       & NPL    & 2017 &   -              & LFSD                   &  Bayesian framework, boundary prior, color/depth-induced contrast               &\ding{51}&\ding{51}&                 &          &                                                   \\
\cline{2-12}
&LFS~\cite{li2017saliency}               & TPAMI  & 2017 &     -            & LFSD                   &  An extension of~\cite{Li2014SaliencyDOLFS}\qquad\qquad\qquad\qquad\qquad\qquad\qquad &\ding{51}&          &                 &          &\ding{51}                                         \\
\cline{2-12}
%\qquad\qquad\qquad\qquad\qquad\qquad\qquad\qquad\qquad\qquad
&MA~\cite{zhang2017MA}                        & TOMM   & 2017 &    -             & LFSD + HFUT-Lytro              &  Superpixels intra-cue distinctiveness, light-field flow, \textbf{new dataset (HFUT-Lytro)}               &\ding{51}&\ding{51}&\ding{51}       &          &                                                   \\
\cline{2-12}
&SDDF~\cite{Wang2018AccurateSSDDF}          & MTAP   & 2018 &    -             &        LFSD          &     Background priors, gradient operator, color contrast, local binary pattern histograms &\ding{51}  & &                 &          &                                                   \\
\cline{2-12}
&SGDC~\cite{Wang2018SalienceGDSGDC}           & CVPR   & 2018 &    -             & LFSD                   &  Focusness cues, color and depth contrast\qquad\qquad\qquad\qquad     &\ding{51}&\ding{51}&                 &          &                                                   \\
\cline{2-12}
&RDFD~\cite{Wang2020RegionbasedDRDFD}         & MTAP   & 2020 &   -              & LFSD                   &  Region-based depth feature  descriptor,  dark channel prior,  multi-layer cellular automata              &\ding{51}&          &                 &          &                                                   \\
\cline{2-12}
&DCA~\cite{Piao2020SaliencyDVDCA}             & TIP    & 2020 &  -           & LFSD                   &  Depth-induced cellular automata, object-guided depth             &\ding{51}&\ding{51}&                 &          &                                                   \\ \hline \hline
& DLLF~\cite{Wang2019DeepLFDLLF}               & ICCV   & 2019 & DUTLF-FS          & LFSD + DUTLF-FS            & VGG-19, attention subnetwork, ConvLSTM,  adversarial examples, \textbf{new dataset (DUTLF-FS)}             &\ding{51}&          &                 &          &            \\  \cline{2-12}
\multirow{7}{*}{\begin{sideways} \textbf{Deep learning models}\end{sideways}}&DLSD~\cite{Piao2019DeepLSDLSD}               & IJCAI  & 2019 & DUTLF-MV          & DUTLF-MV                 & View synthesis network, multi-view detection/attention, VGG-19, \textbf{new dataset (DUTLF-MV)}             &          &          &\ding{51}       &          &\ding{51}                                         \\   \cline{2-12}
&MoLF~\cite{Zhang2019MemoryorientedDFMoLF}    & NIPS   & 2019 & DUTLF-FS          & \tabincell{c}{HFUT-Lytro + LFSD \\+ DUTLF-FS}       & VGG-19, memory-oriented spatial fusion, memory-oriented feature integration               &\ding{51}&          &                 &          &\ding{51}                                          \\  \cline{2-12}
 &ERNet~\cite{Piao2020ExploitARERNet}          & AAAI   & 2020 &  \tabincell{c}{DUTLF-FS \\+ HFUT-Lytro}     & \tabincell{c}{HFUT-Lytro + LFSD \\+ DUTLF-FS}       & VGG-19, ResNet-18, multi-focusness recruiting/screening modules, distillation           &\ding{51}&          &                 &          &\ding{51}                                          \\  \cline{2-12}
&LFNet~\cite{Zhang2020LFNetLFNet}             & TIP    & 2020 & DUTLF-FS          & \tabincell{c}{HFUT-Lytro + LFSD \\+ DUTLF-FS}       & VGG-19, refine unit, attention block, ConvLSTM             &\ding{51}&          &                 &          &          \\   \cline{2-12}
 &MAC~\cite{Zhang2020LightFSLFDCN}           & TIP    & 2020 & Lytro  Illum    & \tabincell{c}{Lytro Illum + LFSD \\+ HFUT-Lytro} &  Micro-lens images/image arrays, DeepLab-v2, model angular changes, \textbf{new dataset (Lytro Illum)}        &          &          &                 &\ding{51}&\ding{51}                                          \\  \cline{2-12}
 &MTCNet~\cite{zhang2020multiMTCNet}           & TCSVT    & 2020 & Lytro  Illum    & \tabincell{c}{Lytro Illum \\+ HFUT-Lytro} &  Edge detection, depth inference, feature-enhanced salient object generator       &          &          &     \ding{51}        & &                                         \\ \hline \hline
\multirow{2}{*}{\begin{sideways}\textbf{Reviews}\end{sideways}} & CS \cite{Zhang2015LightFSCS}                 & NEURO  & 2015 &  -       & LFSD                   &   Comparative study between 2D \emph{vs.} light field saliency        &     &          &                 &          &                         \\   \cline{2-12}
&RGBDS~\cite{Zhou2020RGBDSO}            & CVM  & 2020 &      -          & -                      &    In-depth RGB-D SOD survey, brief review of light field SOD       &          &          &                 &          &                         \\    \hline
\end{tabular}
}
\label{model table}
\end{table*}

\subsubsection{Traditional Models}

As summarized in Table \ref{model table}, traditional light field SOD models often extend various hand-crafted features/hypotheses which are widely adopted in conventional saliency detection \cite{Borji2019SalientOD}, such as global/local color contrast, background priors, and object location cues, to the case of light field data. Some tailored features like focusness, depth, and light-field flow, are incorporated as well. Besides, these traditional models tend to employ some post-refinement steps, \emph{e.g.}, an optimization framework~\cite{Zhang2015SaliencyDILF,Sheng2016RelativeLFRL,zhang2017MA,Wang2018SalienceGDSGDC,Piao2020SaliencyDVDCA} or CRF~\cite{Piao2020SaliencyDVDCA}, to achieve saliency maps with better spatial consistency and more accurate object boundaries. Regarding the data forms leveraged, almost all traditional models work with focal stacks, while depth is incorporated by some of them. Only two traditional models consider using the multi-view~\cite{zhang2017MA} and micro-lens data~\cite{Sheng2016RelativeLFRL}. Further, due to early dataset scarcity, almost all traditional models are evaluated only on the small LFSD dataset constructed by \cite{Li2014SaliencyDOLFS}. Despite early progresses made by these traditional models, due to general limitations of hand-crafted features, they can hardly generalize well to challenging and complex scenarios comparing to modern deep learning models. Here below we briefly review the key features of these traditional models without taxonomy, because they adopt overlapped features but quite diverse computational techniques.

\indent \textbf{LFS}~\cite{Li2014SaliencyDOLFS} was the pioneering and earliest work on light field SOD, where the first dataset was also proposed. LFS first incorporated the focusness measure with location priors to determine the background and foreground slices. Then, in the all-focus\footnote{In this paper, ``all-focus'' and ``all-in-focus'' are used interchangeably. Their meanings are the same.} image, it computed the background prior and contrast cues to detect saliency candidates. Finally, a saliency map was generated by incorporating the saliency candidates in the all-focus image with those in the foreground slices, where objectness cues were used to weight the candidates. An extension of this work was published in~\cite{li2017saliency}.

\textbf{WSC}~\cite{Li2015WSC} was proposed as a unified framework for 2D, 3D and light field SOD problems, which can handle heterogeneous data. Based on the weighted sparse coding framework, the authors first used a non-saliency dictionary to reconstruct a reference image, where patches with high reconstruction error were selected as the saliency dictionary. This saliency dictionary was later refined by iteratively running the weighted sparse framework to achieve the final saliency map. For the light field data, features used for dictionary construction were derived from the all-focus RGB image, depth map, and also focal stacks.

\begin{figure}
\centering
\includegraphics[width=0.48\textwidth]{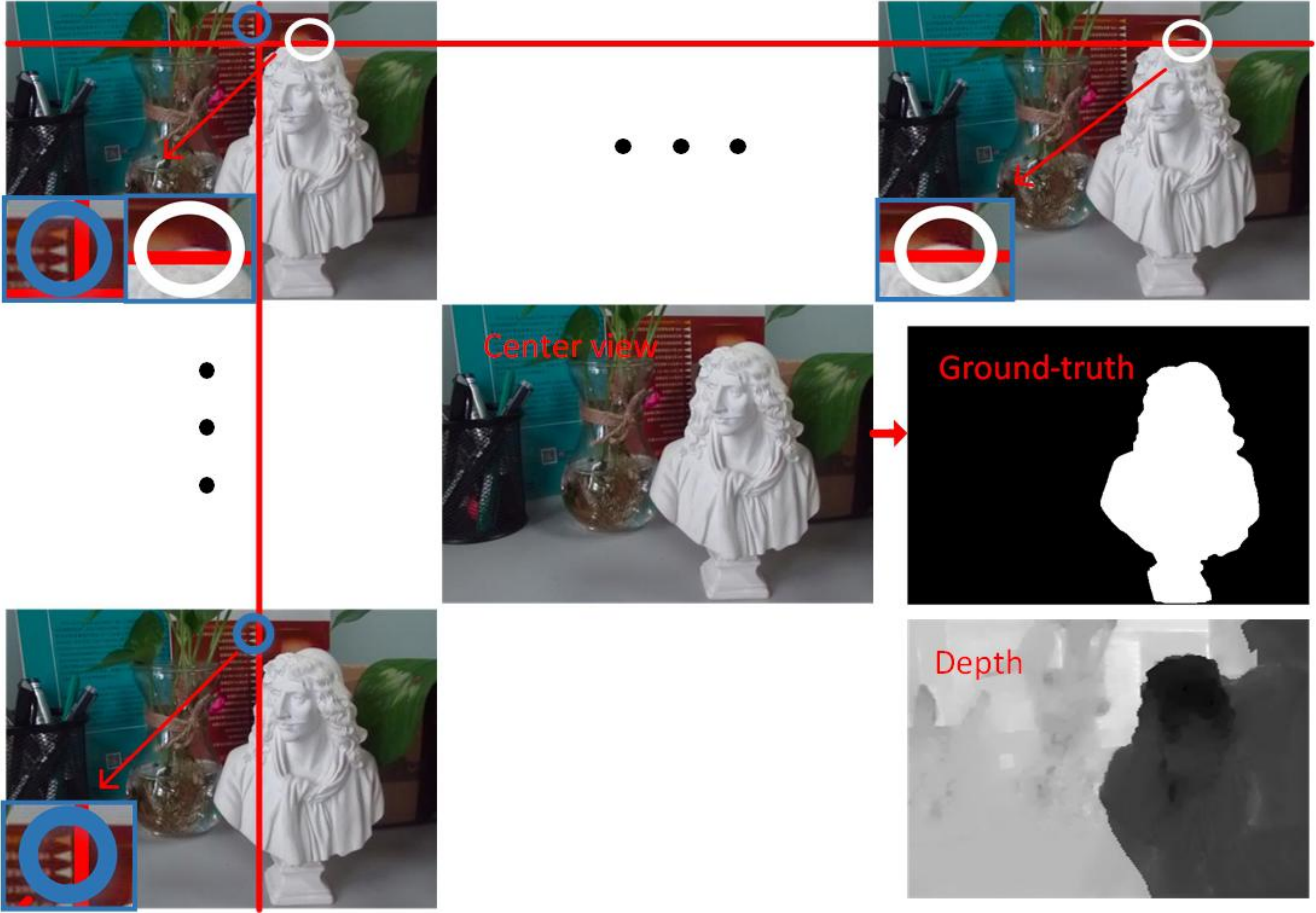}
\caption{Multi-view images (including the center-view image), the depth map, and the ground-truth. Notice the inconspicuous parallax (disparity) conveyed by the multi-view images (zoomed in as the bottom-left in each multi-view image).}
\label{multi-view images, depth and ground truth}
\end{figure}

\textbf{DILF}~\cite{Zhang2015SaliencyDILF} computed the depth-induced contrast saliency and color contrast saliency from the all-focus image and depth image, which were then used to generate a contrast saliency map. It also computed the background priors based on the focusness measure embedded in the focal stacks and used them as weights to eliminate background distraction and enhance saliency estimation.

\textbf{RL}~\cite{Sheng2016RelativeLFRL} proposed to estimate the relative locations of scene points using a filtering process. %It designed two filters, namely the foreground and background filter, to calculate the relative locations in a raw light field image.
Such relative locations, which can be treated as conveying scene depth information, were then incorporated with the robust background detection and saliency optimization framework proposed in~\cite{Zhu2014SaliencyOF} to achieve enhanced saliency detection.

\textbf{BIF}~\cite{Wang2017BIF} used the Bayesian framework to fuse multiple features extracted from RGB images, depth maps and focal stacks. Inspired by image SOD methods, this model utilized a boundary connectivity prior, background likelihood scores and color contrast to generate background probability maps, foreground slices, color-based saliency maps and depth-induced contrast maps, which are fused by a two-stage Bayesian scheme.

\textbf{MA}~\cite{zhang2017MA} measured the saliency of a superpixel by computing the intra-cue distinctiveness between two superpixels, where features considered include color, depth, and flow inherited from different focal planes and multiple viewpoints. The light-field flow was first employed in this method, estimated from focal stacks and multi-view sequences, to capture depth discontinuities/contrast. The saliency measure was later enhanced using a location prior and a random-search-based weighting strategy. In addition, the authors proposed a new light field SOD dataset, which was the largest at that time.

\textbf{SDDF}~\cite{Wang2018AccurateSSDDF} made use of depth information embedded in focal stacks to conduct accurate saliency detection. A background measurement was first obtained by applying a gradient operator to focal stack images, and the focal slice with the highest measurement was chosen as the background layer. A coarse prediction was generated by separating the background and foreground in the all-focus image using the derived background regions, and the final saliency map was calculated globally via both color and texture (local binary pattern histograms) contrast based on the coarse saliency map.

\textbf{SGDC}~\cite{Wang2018SalienceGDSGDC} presented an approach named contrast-enhanced saliency detection for optimizing the multi-layer light field display. It first computed a superpixel-level focusness map for each refocused image and then chose the refocused image with the highest background likelihood score to derive background cues. Such focusness background cues were later incorporated with color and depth contrast saliency. The final results were optimized by the optimization framework proposed in~\cite{Zhu2014SaliencyOF}.

\textbf{RDFD}~\cite{Wang2020RegionbasedDRDFD} addressed the light field SOD problem via a multiple cue integration framework. A region-based depth feature descriptor (RDFD) defined over the focal stack was proposed, which was based on the observation that dark channel priors \cite{he2010single} can be used to estimate the degree of defocusing/blurriness. The RDFD was generated by integrating the degrees of defocusing over all focal stack images, alleviating the limitation of requiring accurate depth maps. RDFD features were used to compute a region-based depth contrast map and a 3D spatial distribution prior. These cues were merged into a single map using a multi-layer cellular automata (MCA).

\textbf{DCA}~\cite{Piao2020SaliencyDVDCA} proposed a depth-induced cellular automata (DCA) for light field SOD. Firstly, it used the focusness and depth cue to calculate the object-guided depth map and select background seeds. Based on the seeds, a contrast saliency map was computed and multiplied with the object-guided depth map to achieve a depth-induced saliency map, which was subsequently optimized by DCA. Finally, the optimized map was combined with the depth-induced saliency map. A Bayesian fusion strategy and CRF were employed to refine the prediction.

\begin{figure*}[!ht]
\centering
\includegraphics[width=0.98\textwidth]{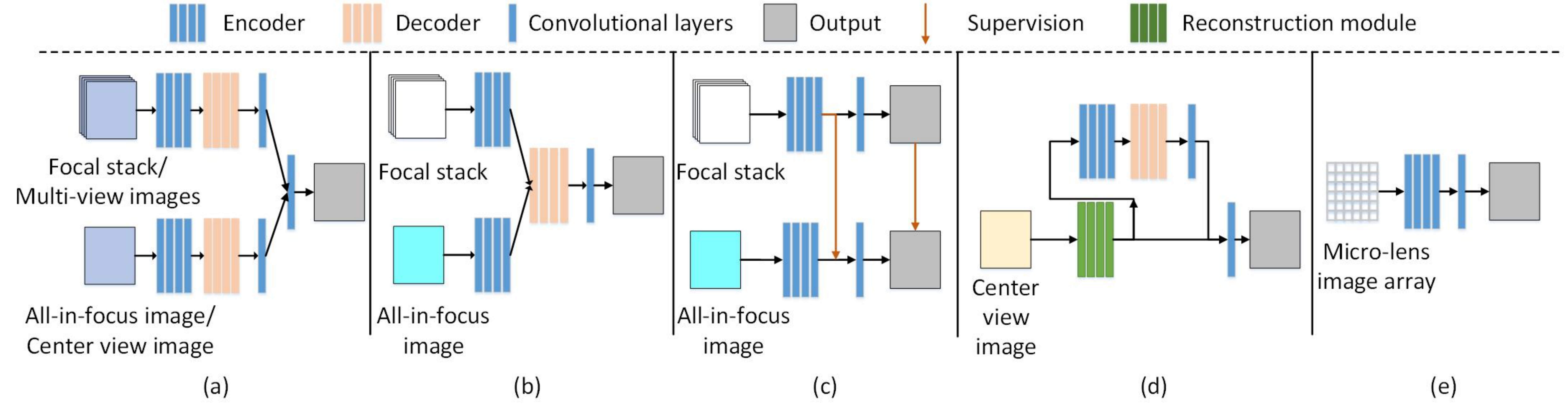}
\caption{Architectures of existing deep light field SOD models. (a) Late-fusion (DLLF \cite{Wang2019DeepLFDLLF}, MTCNet~\cite{zhang2020multiMTCNet}). (b) Middle-fusion (MoLF \cite{Zhang2019MemoryorientedDFMoLF}, LFNet \cite{Zhang2020LFNetLFNet}). (c) Knowledge distillation-based (ERNet \cite{Piao2020ExploitARERNet}). (d) Reconstruction-based (DLSD \cite{Piao2019DeepLSDLSD}). (e) Single-stream (MAC \cite{Zhang2020LightFSLFDCN}). Note that (a) utilizes the focal stack/multi-view images and all-focus/center-view image, while (b)-(c) utilize the focal stack and all-focus image. (d)-(e) utilize the center-view image and micro-lens image array.}
\label{Category of previous deep SOD models}
\end{figure*}

\subsubsection{Deep Learning-based Models}

Due to the powerful learning ability of deep neural networks, deep learning-based models can achieve superior accuracy and performance \cite{Wang2019DeepLFDLLF} over traditional light field SOD models. Another advantage of deep models comparing to the latter is that they can directly learn from a large amount of data without hand-crafted feature engineering. Therefore, as overviewed in Table \ref{model table}, one can see that the scarcity of datasets has been somewhat alleviated in the deep learning era, as three new datasets have been introduced to better train deep neural networks. Still, most deep models take a focal stack as network input. Due to the multi-variable property of focal stacks, modules such as attention mechanisms~\cite{Wang2019DeepLFDLLF,Piao2019DeepLSDLSD,Zhang2019MemoryorientedDFMoLF,Zhang2020LFNetLFNet,Piao2020ExploitARERNet} and ConvLSTMs~\cite{Wang2019DeepLFDLLF,Zhang2019MemoryorientedDFMoLF,Zhang2020LFNetLFNet,Piao2020ExploitARERNet} are preferred.
We argue that there may be different ways of taxonomy for deep models. A straightforward way is to categorize according to what kind of light field data is utilized, as indicated in Table \ref{model table}. While there are four models, namely DLLF \cite{Wang2019DeepLFDLLF}, MoLF \cite{Zhang2019MemoryorientedDFMoLF}, ERNet \cite{Piao2020ExploitARERNet}, LFNet \cite{Zhang2020LFNetLFNet} resorting to focal stacks, DLSD \cite{Piao2019DeepLSDLSD} and MTCNet \cite{zhang2020multiMTCNet} utilize multi-view images, and MAC \cite{Zhang2020LightFSLFDCN} explores micro-lens images. Different input data forms often lead to different network designs. Note that for DLSD \cite{Piao2019DeepLSDLSD}, multi-view images processed are indeed rendered from an input single-view image, so this method can be applied to cases no matter multi-view images are available or not.

However, since using deep learning-based techniques for light field SOD is the leading trend, in this paper, we further propose to divide existing deep models into five categories according to their architectures, including \textbf{late-fusion scheme},  \textbf{middle-fusion scheme}, \textbf{knowledge distillation-based scheme}, \textbf{reconstruction-based scheme}, and \textbf{single-stream scheme}, as illustrated in Fig. \ref{Category of previous deep SOD models}. Their descriptions and associated models are briefly introduced as follows.

\textbf{Late-fusion.} Late-fusion models (Fig. \ref{Category of previous deep SOD models} (a), DLLF \cite{Wang2019DeepLFDLLF}, MTCNet~\cite{zhang2020multiMTCNet}) aim at obtaining individual predictions from the input focal stack/multi-view images and all-focus/center-view image, and then simply fuses the results. Note that late fusion is a classical strategy also widely adopted in previous multi-modal detection works (\emph{e.g.}, RGB-D SOD \cite{Zhou2020RGBDSO}, RGB-D semantic segmentation \cite{Shelhamer2017FullyCN,Fu2021siamese}) due to its simplicity and easy implementation. However, the fusion process is restrained to the last step with relatively simple integrative computation.

$\bullet$ \textbf{DLLF} \cite{Wang2019DeepLFDLLF} adopted a two-stream fusion framework that explored focal stacks and all-in-focus images separately. In the focal stack stream, DLLF first extracted features from cascaded focal slices through a fully convolutional network. Diverse features from different slices were then integrated by a recurrent attention network, which employed an attention subnetwork and ConvLSTM~\cite{shi2015convolutional} to adaptively incorporate weighted features of slices and exploit their spatial relevance. The generated map was lately combined with another saliency map derived from the all-in-focus image. In addition, to address the limitation of data for training deep networks, a new large dataset was introduced.

$\bullet$ \textbf{MTCNet}~\cite{zhang2020multiMTCNet} proposed a two-stream multi-task collaborative network, consisting of a saliency-aware feature aggregation module (SAFA) and multi-view inspired depth saliency feature extraction (MVI-DSF) module, to extract representative saliency features with the aid of the correlation mechanisms across edge detection, depth inference and salient object detection. SAFA simultaneously extracted focal-plane, edge and heuristic saliency features from a center-view image, while MVI-DSF inferred depth saliency features from a set of multi-view images. Finally, MTCNet combined the extracted features using a feature-enhanced operation to obtain the final saliency map.

\textbf{Middle-fusion.} The middle-fusion strategy (Fig. \ref{Category of previous deep SOD models} (b), MoLF \cite{Zhang2019MemoryorientedDFMoLF}, LFNet \cite{Zhang2020LFNetLFNet}) extracts features from the focal stack and all-focus image in a two-stream manner. Fusion across intermediate features is then done by an elaborate and complex decoder. Comparing to the late-fusion strategy in Fig. \ref{Category of previous deep SOD models} (a), the main differences are that the features fused are usually hierarchical and intermediate, and the decoder is also a relatively deep convolutional network to mine more complex integration rules.

$\bullet$  \textbf{MoLF}~\cite{Zhang2019MemoryorientedDFMoLF} was featured by  a memory-oriented decoder that consists of spatial fusion module (Mo-SFM) and feature integration module (Mo-FIM), in order to resemble the memory mechanism of how humans fuse information. Mo-FSM utilized an attention mechanism to learn the importance of different feature maps and a ConvLSTM~\cite{shi2015convolutional} to gradually refine spatial information. In Mo-FIM, a scene context integration module (SCIM) and ConvLSTM were employed to learn channel attention maps and summarize spatial information.

$\bullet$ \textbf{LFNet}~\cite{Zhang2020LFNetLFNet} proposed a two-stream fusion network to refine complementary information and integrate focusness and blurriness, which change gradually in focal slices. Features extracted from the all-focus image and focal stack are fed to a light field refinement module (LFRM) and integration module (LFIM) to generate a final saliency map. In LFRM, features extracted from a single  slice were fed to a refinement unit to learn the residuals. In LFIM, an attention block was used to adaptively weight and aggregate slice features.

\textbf{Knowledge distillation-based.} Different from the aforementioned late-fusion/middle-fusion schemes that investigate how to fuse focal stack and all-focus features, the knowledge distillation-based scheme (Fig. \ref{Category of previous deep SOD models} (c), ERNet \cite{Piao2020ExploitARERNet}) attempts to transfer focusness knowledge of a teacher network that handles focal stacks, to a student network that processes all-focus images. It uses both the features and prediction from the focal stack stream to supervise those features and prediction obtained from the all-focus stream, effectively boosting the performance of the latter. In this sense, the student network is actually an RGB SOD network augmented by extra light field knowledge during training.

$\bullet$ \textbf{ERNet}~\cite{Piao2020ExploitARERNet} consisted of two-stream teacher-student networks based on knowledge distillation. The teacher network used a multi-focusness recruiting module (MFRM) and a multi-focusness screening module (MFSM) to recruit and distil knowledge from focal slices, while the student network took a single RGB image as input for computational efficiency and was enforced to hallucinate multi-focusness features as well as the prediction from the teacher network.

\textbf{Reconstruction-based.} Comparing to the above three categories, the reconstruction-based scheme (Fig. \ref{Category of previous deep SOD models} (d), DLSD \cite{Piao2019DeepLSDLSD}) focuses on a different aspect as well, namely reconstructing light field data/information from a single input image. This is indeed another interesting topic as light field has various data forms (Section \ref{Forms of Light Field Data}). With the assistance of the reconstructed light field, an encoder-decoder architecture with middle-/late-fusion strategy can then be employed to complete light field SOD. In other words, this scheme is similar to the role of student network in the knowledge distillation-based scheme, namely that it is essentially an RGB SOD network augmented by extra light field knowledge during training (in this case, learns to reconstruct light field data).

$\bullet$  \textbf{DLSD}~\cite{Piao2019DeepLSDLSD} treated light field SOD as two sub-problems: light field synthesis from a single-view image and light-field-driven SOD. This model first employed a light field synthesis network, which estimated depth maps along the horizontal and vertical directions with two independent convolutional networks. According to the depth maps, the single-view image was warped into horizontal and vertical viewpoints of the light field. A light-field-driven SOD network, consisting of a multi-view saliency detection subnetwork and multi-view attention module, was designed for saliency prediction. Specifically, this model inferred a saliency map from a 2D single-view image, but utilized the light field (the multi-view data) as a middle bridge. To train the model, a new dataset containing multi-view images and pixel-wise ground-truth of central view was introduced.

\begin{table*}[!htb]
\centering
\caption{Overview of light field SOD datasets. We use the following abbreviations: MOP=multiple-object proportion (the percentage of images in the entire dataset that have more than one object per image), FS=focal stacks, DE=depth maps, MV=multi-view images, ML=micro-lens images, GT=ground-truth, Raw=raw light field data. Here, FS, MV, DE, ML, GT and Raw indicate the data forms provided by the datasets.}
\resizebox{\textwidth}{1.2cm}{
\begin{tabular}{l|c|c|c|c|cccccc|c}
\hline
Dataset                               & Scale & Spatial resolution & Angular resolution   & MOP  & FS  & MV  & DE  & ML & GT  & Raw   & Device            \\ \hline\hline
LFSD~\cite{Li2014SaliencyDOLFS}         & 100 (No offical split)   & 360 $\times$ 360     & -       & 0.04  & \ding{51}   &             & \ding{51}  &                   & \ding{51}    &\ding{51}  & Lytro        \\  \hline
HFUT-Lytro~\cite{zhang2017MA}                 & 255 (No offical split)   & 328 $\times$ 328     & 7 $\times$ 7       & 0.29     & \ding{51}   & \ding{51}  & \ding{51}  &                   & \ding{51}    &           & Lytro        \\ \hline
DUTLF-FS~\cite{Wang2019DeepLFDLLF}        & 1462 (1000 train,  462 test)  & 600 $\times$ 400     & -       & 0.05     & \ding{51}   &             &   \ding{51}   &      & \ding{51}   &       & Lytro Illum  \\ \hline
DUTLF-MV~\cite{Piao2019DeepLSDLSD}        & 1580 (1100 train,  480 test)  & 590 $\times$ 400     & 7 $\times$ 7       & 0.04     &              & \ding{51}  &           &            & \ding{51}   &          & Lytro Illum  \\ \hline
Lytro Illum~\cite{Zhang2020LightFSLFDCN}& 640 (No offical split)   & 540 $\times$ 375     & 9 $\times$ 9       & 0.15     &              &             &           & \ding{51}        & \ding{51}    &       \ding{51}     & Lytro Illum  \\ \hline
\end{tabular}}
\label{dataset table}
\end{table*}

\textbf{Single-stream.} Lastly, the single-stream scheme (Fig. \ref{Category of previous deep SOD models} (e), MAC \cite{Zhang2020LightFSLFDCN}) is inspired by the fact that light field can be formulated in a single image representation, namely the micro-lens image array \cite{Zhang2020LightFSLFDCN}. Therefore, unlike Fig. \ref{Category of previous deep SOD models} (a) and (b), this scheme processes the micro-lens image array directly using a single bottom-up stream, without explicit feature fusion.

$\bullet$ \textbf{MAC}~\cite{Zhang2020LightFSLFDCN} was an end-to-end deep convolutional network for light field SOD with micro-lens image arrays as input. Firstly, it adopted a MAC (Model Angular Changes) block tailored to model angular changes in individual local micro-lens image and then fed the extracted features to a modified DeepLab-v2 network \cite{Chen2018DeepLabSI}, capturing multiscale information and long-range spatial dependencies. Together with the model, a new Lytro Illum dataset containing high-quality micro-lens image arrays was proposed.

\subsubsection{Other Review Works}

\indent \textbf{CS}~\cite{Zhang2015LightFSCS}.
This paper provided a comparative study between light field saliency and 2D saliency, showing the advantage of conducting the SOD task on light field data over single 2D images. It compared the classical model LFS~\cite{Li2014SaliencyDOLFS} with eight 2D saliency models on the LFSD dataset \cite{Li2014SaliencyDOLFS}. Five evaluation metrics were used in the paper to show that the light field saliency model achieved better and more robust performance than conventional 2D models.

\textbf{RGBDS}~\cite{Zhou2020RGBDSO}.
This paper conducted an in-depth and comprehensive survey on RGB-D salient object detection. It reviewed existing RGB-D SOD models from various perspectives, as well as the related benchmark datasets in detail. Considering the fact that light fields can also provide depth maps, the authors also briefly reviewed the light field SOD models and datasets. However, because the main focus of this paper was on RGB-D SOD, very few contents are given to review light field SOD, and no associated benchmarking was conducted.

\subsection{Light Field SOD Datasets}
\label{Light field SOD datasets}

At present, there are five datasets introduced for the light field SOD task, including LFSD~\cite{Li2014SaliencyDOLFS}, HFUT-Lytro~\cite{zhang2017MA}, DUTLF-FS~\cite{Wang2019DeepLFDLLF}, DUTLF-MV~\cite{Piao2019DeepLSDLSD}, and Lytro Illum~\cite{Zhang2020LightFSLFDCN}. Beyond Table \ref{model table}, we summarize details of these datasets in Table \ref{dataset table} and show some samples from four datasets (\ie, LFSD, HFUT-Lytro, Lytro Illum, and DUTLF-FS) in Fig. \ref{rgb, depth and GT}. A brief introduction is given as follows:

\begin{figure*}[!ht]
\centering
\includegraphics[width=0.98\textwidth]{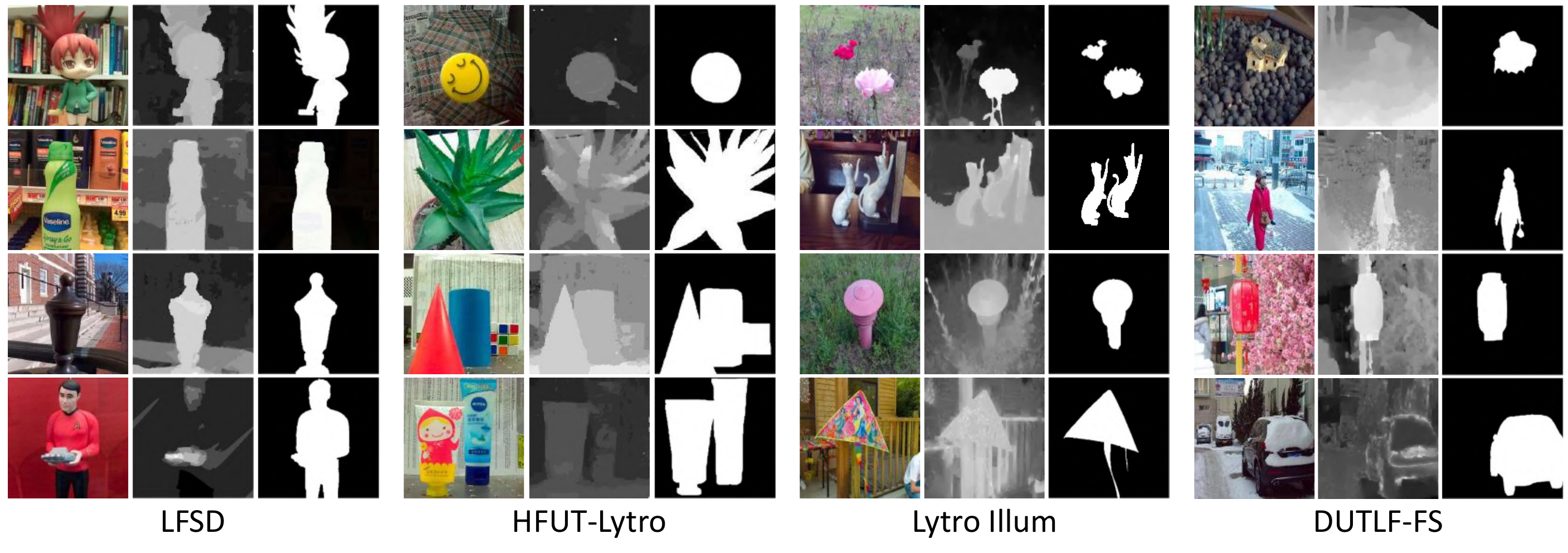}\vspace{-0.3cm}
\caption{Examples of RGB images, depth maps, and ground-truths (GTs) from four datasets: LFSD~\cite{Li2014SaliencyDOLFS}, HFUT-Lytro~\cite{zhang2017MA}, Lytro Illum~\cite{Zhang2020LightFSLFDCN} and DUTLF-FS~\cite{Wang2019DeepLFDLLF}. In each group, RGB images, depth maps and GTs are shown from left to right.}
\label{rgb, depth and GT}
\end{figure*}

\textbf{LFSD}\footnote{\url{https://sites.duke.edu/nianyi/publication/saliency-detection-on-light-field/}}~\cite{Li2014SaliencyDOLFS} was the first light field dataset collected for SOD, and contains 60 indoor and 40 outdoor scenes. This dataset was captured by a Lytro camera and provides a focal stack, all-in-focus image, depth map and the corresponding ground-truth for each light field. The image spatial resolution is 360$\times$360. Besides, raw light field data is also available in LFSD. Note that most images in this dataset contain one single center-placed object with a relatively simple background.

\textbf{HFUT-Lytro}\footnote{\url{https://github.com/pencilzhang/MAC-light-field-saliency-net}\label{footnote:MAC}}~\cite{zhang2017MA} contains 255 light fields for both indoor and outdoor scenes. Each light field contains a focal stack whose slice numbers vary from 1 to 12. The angular resolution is 7 $\times$ 7 and the spatial resolution is 328 $\times$ 328. Notably, focal stacks, all-in-focus images, multi-view images and coarse depth maps are all provided in this dataset.  Several challenges regarding SOD, \eg, occlusions, cluttered background and appearance changes, are present in HFUT-Lytro.

\textbf{DUTLF-FS}\footnote{\url{https://github.com/OIPLab-DUT/ICCV2019\_Deeplightfield\_Saliency}}~\cite{Wang2019DeepLFDLLF} is one of the largest light field SOD datasets to date,
containing 1,462 light fields in total. It was acquired by a Lytro Illum camera in both indoor and outdoor scenes. The entire dataset is officially divided into 1,000 training samples and 462 testing samples. All-focus images, focal stacks and the corresponding ground-truth are provided for different light fields. The slice number of a focal stack ranges from 2 to 12, and the image spatial resolution is 600 $\times$ 400. It is worth noting that DUTLF-FS covers various challenges, including different types of objects, low appearance contrast between salient objects and their background, and varied object locations.

\textbf{DUTLF-MV}\footnote{\url{https://github.com/OIPLab-DUT/IJCAI2019-Deep-Light-Field-Driven-Saliency-Detection-from-A-Single-View}}~\cite{Piao2019DeepLSDLSD} is another large-scale light field dataset for SOD, which was generated from the same database as DUTLF-FS (with 1,081 identical scenes). In contrast to other datasets, this dataset was proposed to better exploit the angular cues. Therefore, only multi-view images with respect to horizontal and vertical viewpoints are available, together with the ground-truth of the center view image. DUTLF-MV contains 1,580 light fields in total, and is officially divided into training and test sets with 1,100 and 480 samples, respectively. The spatial resolution of each image is 400 $\times$ 590 and the angular resolution is 7 $\times$ 7.

\textbf{Lytro Illum}\textsuperscript{\ref{footnote:MAC}}~\cite{Zhang2020LightFSLFDCN} contains 640 high-quality light fields captured by a Lytro Illum camera. The images in this dataset vary significantly in object size, texture, background clutter and illumination. Lytro Illum provides center-view images, micro-lens image arrays, raw light field data, as well as the corresponding ground-truths of the center-view images. The resolution of micro-lens image arrays is 4860 $\times$ 3375, while center-view images and ground-truths have a 540 $\times$ 375 spatial resolution. The angular resolution can be inferred as 9 $\times$ 9.

\begin{figure*}
\centering
\includegraphics[width=0.98\textwidth]{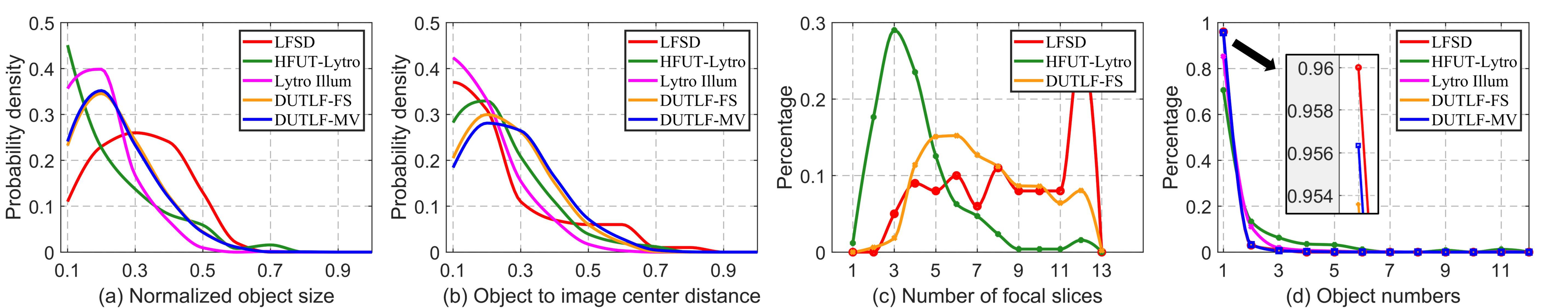}
\caption{Statistics of light field datasets, including LFSD \cite{Li2014SaliencyDOLFS}, HFUT-Lytro \cite{zhang2017MA}, Lytro Illum \cite{Zhang2020LightFSLFDCN}, DUTLF-FS \cite{Wang2019DeepLFDLLF} and DUTLF-MV \cite{Piao2019DeepLSDLSD}. From left to right: (a) distribution of the normalized object size, (b) distribution of the normalized distance between the object and image center, (c) statistics on focal slice numbers, and (d) statistic on object numbers.}
\label{statistics of all dataset}
\end{figure*}

\begin{figure}
\centering
\includegraphics[width=0.48\textwidth]{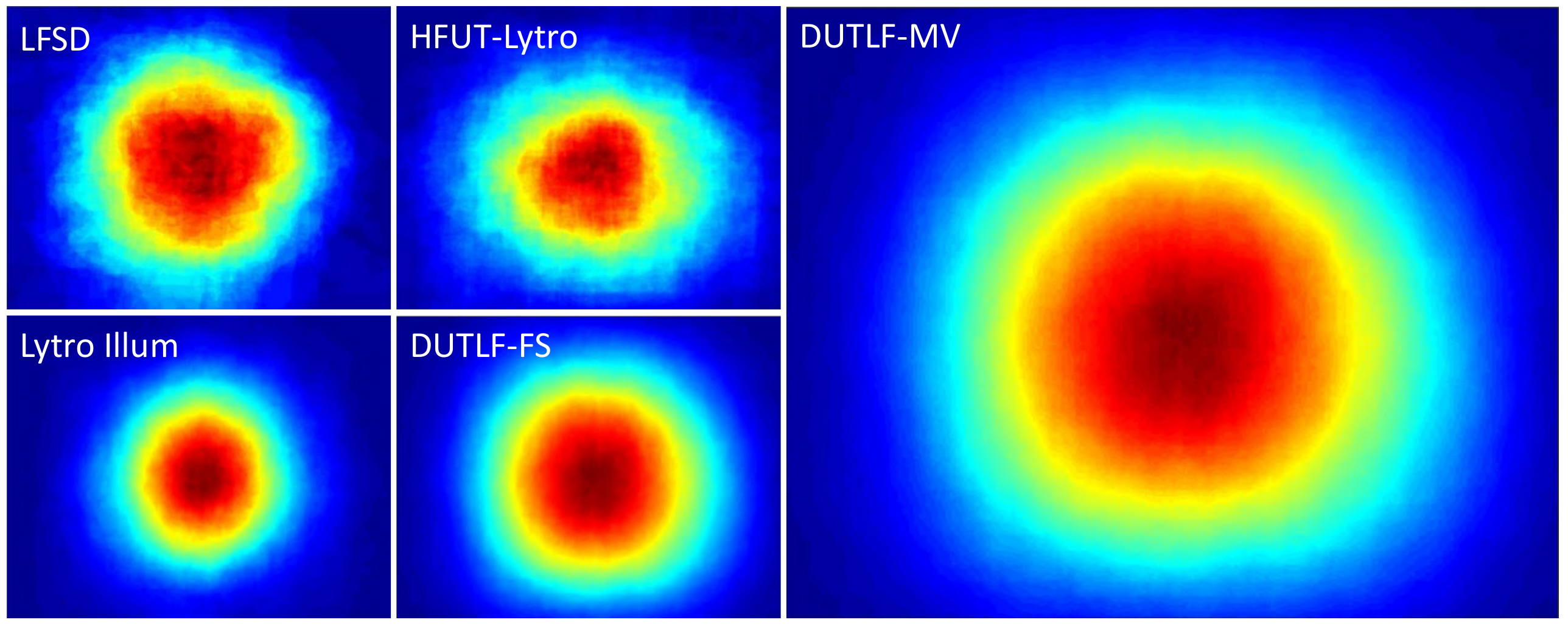}
\caption{Object location distribution maps of five datasets (warmer color means higher probability). The maps are computed by averaging ground-truth masks.}
\label{Average ground-truth}
\end{figure}

\textbf{Dataset Analysis}.  From the summarization in Table \ref{dataset table}, we can observe two issues existed in current datasets, namely scale limitation and non-unified data forms. Compared to the large datasets constructed for the conventional SOD task, such as DUT-OMRON (5,168 images)~\cite{Yang2013SaliencyDV}, MSRA10K (10,000 images)~\cite{Cheng2011GlobalCB} and DUTS (15,572 images)~\cite{Wang2017LearningTD}, the existing light field SOD datasets are still small, making it somewhat difficult to evaluate data-driven models and train deep networks. Besides, their data forms are not always consistent. For example, Lytro Illum does not provide focal stacks, while DUTLF-FS/DUTLF-MV only provide focal stacks/multi-view images without offering the raw data. This makes comprehensive benchmarking very difficult, because a model using focal stacks as input cannot run on DUTLF-MV and Lytro Illum. We will show how we alleviate this problem in Section \ref{Dataset completion}, and discuss the underlying future directions in Section \ref{challenge and open direction}.

To better understand the above-mentioned datasets, we have conducted statistical analyses, including size ratios of salient objects, distributions of normalized object distances from image centers, numbers of focal slices and numbers of objects. The quantitative results are shown in Fig. \ref{statistics of all dataset} and Fig. \ref{Average ground-truth}. From Fig. \ref{statistics of all dataset} (a), we can see that most objects from these datasets have size ratios lower than 0.6. HFUT-Lytro and Lytro Illum have relatively small objects, while LFSD has objects that are relatively large in size. Fig. \ref{statistics of all dataset} (b) and Fig. \ref{Average ground-truth} clearly show the spatial distributions of objects. From Fig. \ref{Average ground-truth}, we can see that all five datasets present strong center bias, and Fig. \ref{statistics of all dataset} (b) reveals that objects from Lytro Illum are generally the closest to the image centers (also indicated by Fig. \ref{Average ground-truth}).

In addition, the statistics on focal slice numbers are given in Fig. \ref{statistics of all dataset} (c). Note that only three datasets, namely LFSD, HFUT-Lytro, and DUTLF-FS, provide focal slices. The numbers of slices vary from 1 to 12 and there are notable differences between different datasets. The slice numbers corresponding to the distribution peaks on LFSD, HFUT-Lytro and DUTLF-FS are 12, 3, and 6, respectively.
% This is because LFSD provides richer depth information than the other datasets. Besides, a
All three datasets have various slices numbers, indicating that a light field SOD model using focal stacks should be able to handle various numbers of input slices. Lastly, from Fig. \ref{statistics of all dataset} (d), we can see that most images from these datasets have a single object. Also, HFUT-Lytro and Lytro Illum have some images with multiple objects (with higher ``MOP'' in Table \ref{dataset table}), which could be useful for validating models on detecting multiple objects.

\section{Model Evaluation and Benchmark}\label{sec:model_eval_benchmarks}

In this section, we first review five popular evaluation metrics, and then provide a pipeline for achieving dataset unification. Moreover, we carry out a benchmarking evaluation and provide an experimental result analysis.

\subsection{Evaluation Metrics}

In our benchmarking of light field SOD models, we employ nine metrics, which are universally agreed upon and are described as follows:

\textbf{Precision-recall (PR)}~\cite{Achanta2009FrequencytunedSR,Borji2015SalientOD,Cheng2011GlobalCB} curve is defined as:
\begin{align}
    Precision(T) = \frac{|M^T\cap G|}{|M^T |},
    Recall(T) = \frac{|M^T\cap G|}{|G|},
\end{align}
where $M^T$ is a binary mask obtained by thresholding the saliency map with threshold $T$, and $|\cdot|$ is the total area of the mask. $G$ denotes the ground-truth. A comprehensive precision-recall curve is obtained by changing $T$ from 0 to 255.

\textbf{F-measure} $\bm{(F_\beta)}$~\cite{Achanta2009FrequencytunedSR,Borji2015SalientOD,Cheng2011GlobalCB} is defined as the harmonic-mean of precision and recall:
\begin{align}
    F_\beta =\frac{(1+\beta^2 )Precision \cdot Recall}{\beta^2 \cdot Precision+Recall},
\end{align}
where $\beta$ is the weight between $Precision$ and $Recall$, and $\beta^2$ is often set to 0.3 to emphasize more on precision. Since different F-measure scores can be obtained according to different precision-recall pairs, in this paper, we report the maximum F-measure ($F_{\beta}^{\textrm{max}}$) and mean F-measure ($F_{\beta}^{\textrm{mean}}$) computed from the PR curve. Besides, we also report the adaptive F-measure ($F_{\beta}^{\textrm{adp}}$)~\cite{Achanta2009FrequencytunedSR}, whose threshold is computed as the twice of the mean of a saliency map.

\textbf{Mean Absolute Error ($M$)}~\cite{Perazzi2012SaliencyFC} is defined as:
\begin{align}
    M = \frac{1}{N}\sum_{i=1}^N|S_i-G_i|,
\end{align}
where $S_i$ and $G_i$ denote the values at the $i$-th pixel in the saliency map and ground-truth map. $N$ is the total number of pixels in the both map.

\textbf{S-measure} $\bm{(S_\alpha)}$~\cite{Fan2017StructureMeasureAN,Zhao2019ContrastPA} was proposed to measure the spatial structure similarities between the saliency map and ground-truth. It is defined as:
\begin{align}
    S_\alpha = \alpha*S_o+(1-\alpha)*S_r,
\end{align}
where $S_o$ and $S_r$ denote the object-aware and region-aware structural similarity, respectively, and $\alpha$  balances $S_o$ and $S_r$. In this paper, we set $\alpha = 0.5$, as recommended in~\cite{Fan2017StructureMeasureAN}.

 \begin{figure*}
 \centering
 \includegraphics[width=0.88\textwidth]{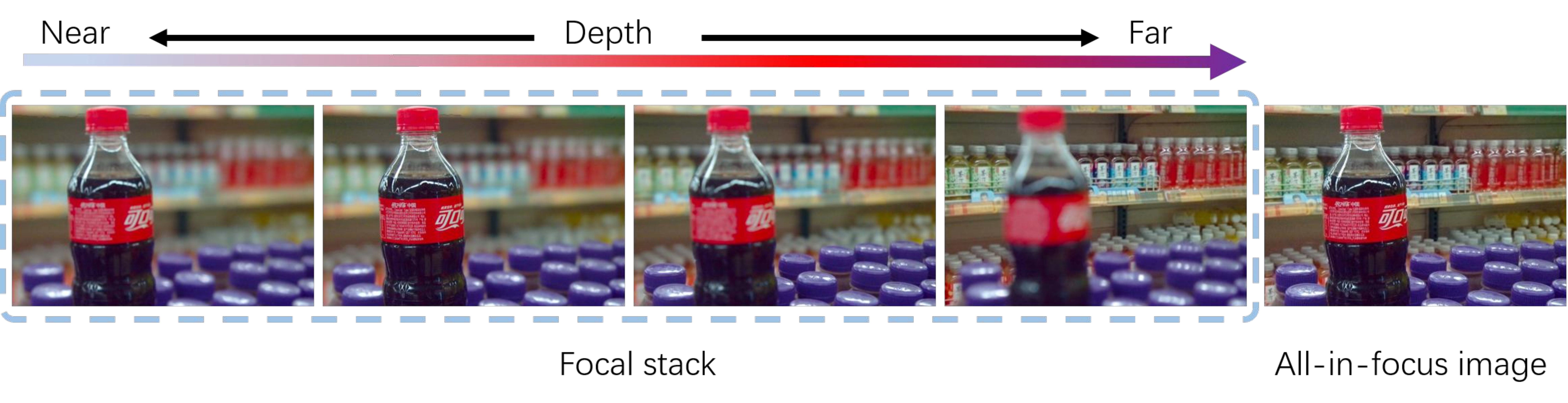}\vspace{-0.3cm}
 \caption{Example of generated focal slices for Lytro Illum~\cite{Zhang2020LightFSLFDCN}, together with the synthesized all-in-focus image.}
 \label{Generated focal stack}
 \end{figure*}

\textbf{E-measure} $\bm{(E_\phi)}$~\cite{Fan2018EnhancedalignmentMF} is a recently proposed metric which considers both the local and global similarity between the prediction and ground-truth. It is defined as:
\begin{align}
    E_\phi = \frac{1}{w*h}\sum_{i=1}^w\sum_{j=1}^h\phi(i,j),
\end{align}
where $\phi(\cdot)$ denotes the enhanced alignment matrix~\cite{Fan2018EnhancedalignmentMF}. $w$ and $h$ are the width and height of the ground-truth map, while $(i,j)$ are pixel indexes. Since $E_\phi$ also performs a comparison between two binary maps, we treat it similarly to the F-measure, thresholding a saliency map with all possible values and reporting the maximum and mean $E_\phi$, denoted as $E_{\phi}^{\textrm{max}}$ and $E_{\phi}^{\textrm{mean}}$. Besides, adaptive $E_\phi$, namely $E_{\phi}^{\textrm{adp}}$, is computed similarly to the adaptive F-measure mentioned above, where the thresholds are two times the mean saliency values~\cite{Achanta2009FrequencytunedSR}.

Note that, higher PR curves, $F_\beta$, $S_\alpha$ and $E_\phi$, and lower $M$ indicate better performance.

\begin{table}[!htbp]
\centering
\renewcommand{\arraystretch}{0.8}
\renewcommand{\tabcolsep}{1.6mm}
\caption{Dataset unification for light field SOD, which is in contrast to Table \ref{dataset table}. About the abbreviations: FS=focal stacks, DE=depth maps, MV=multi-view images, mL=Micro-lens images, Raw=raw light field data. Note that ``\CircPipe'' indicates data that we have completed.}
\begin{tabular}{l|p{0.8cm}<{\centering}|p{0.8cm}<{\centering}|p{0.8cm}<{\centering}|p{0.8cm}<{\centering}|p{0.8cm}<{\centering}}
\hline
Datasets                                 & FS  & MV  & DE  & ML & Raw \\ \hline\hline
LFSD~\cite{Li2014SaliencyDOLFS}        &  \ding{51}  &  \CircPipe   & \ding{51}  &   \CircPipe     &\ding{51}    \\  \hline
HFUT-Lytro~\cite{zhang2017MA}                  & \ding{51}   & \ding{51}  & \ding{51}  &  \CircPipe                  \\ \hline
DUTLF-FS~\cite{Wang2019DeepLFDLLF}           &  \ding{51}  &             &   \ding{51}  &       \\ \hline
DUTLF-MV~\cite{Piao2019DeepLSDLSD}         &     & \ding{51}  &    &       &             \\ \hline
Lytro Illum~\cite{Zhang2020LightFSLFDCN} &   \CircPipe            &   \CircPipe         &   \CircPipe       &  \ding{51} & \ding{51}\\ \hline
\end{tabular}
\label{completion table}
\end{table}

\subsection{Dataset Unification}
\label{Dataset completion}

As shown in Section \ref{Light field SOD datasets}and Table \ref{dataset table}, existing light field SOD datasets face the limitation of having non-unified data forms. This makes comprehensive benchmarking difficult. Since, due to the lack of specific data, some models cannot be correctly evaluated on certain datasets. To alleviate this issue, we generate supplemental data for existing datasets, making them complete and unified. The completed data forms are illustrated in Table \ref{completion table}, marked by ``\CircPipe''. Furthermore, we will release this data on our project site: \url{https://github.com/kerenfu/LFSOD-Survey}
%\url{https://dpfan.net/LFSOD-Survey}
to facilitate future research in this field.

\begin{figure}[!ht]
 \centering
 \includegraphics[width=0.48\textwidth]{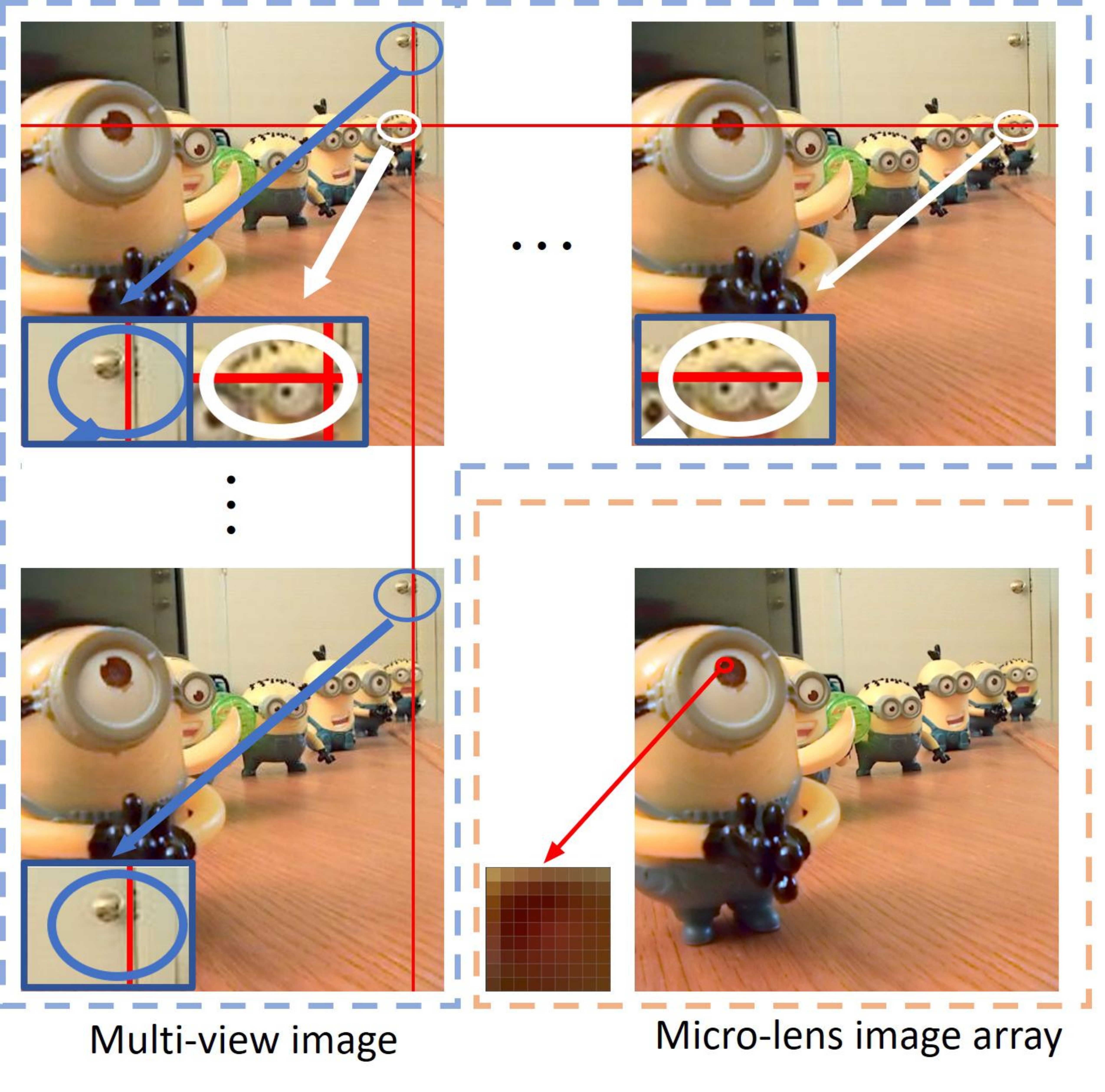}\vspace{-0.3cm}
 \caption{Examples of generated multi-view images ($360 \times 360$) and a micro-lens image array ($1080 \times 1080$) for LFSD dataset \cite{Li2014SaliencyDOLFS}. As in Fig. \ref{multi-view images, depth and ground truth}, the bottom-left of each image shows zoomed-in details to better reflect the parallax. The micro-lens image array is composed of many micro-lens images \cite{Zhang2020LightFSLFDCN}.}
 \label{Generated micro-lens image and multi-view images of LFSD}
\end{figure}

Generally, we can synthesize various data forms using the raw light field data, provided by two datasets, \emph{i.e.}, LFSD and Lytro Illum. For Lytro Illum, we generate focal stacks (including all-in-focus images) and depth maps using the Lytro Desktop software. Regarding focal stack generation, we estimate the approximate focus range for each image scene, and then sample the focal slices within the focus range by an equal step. All-blurred or duplicated slices are removed. The final number of generated focal slices for Lytro Illum ranges from 2 to 16 for each scene, and about 74\% of scenes have more than 6 slices. Fig. \ref{Generated focal stack} shows an example of the generated focal stack.
As mentioned in Section \ref{Forms of Light Field Data}, multi-view images and micro-lens image arrays are generated by sampling the light field data in angular and spatial resolution, respectively. Thus, these two data forms can be transformed between each other. In this way, we generate multi-view images for Lytro Illum from its micro-lens image arrays. We can also synthesize micro-lens image arrays for HFUT-Lytro through the reverse operation. However, we cannot synthesize micro-lens image arrays for DUTLF-MV since the authors have only released the multi-view images in the vertical/horizontal direction.
By using the raw data, we complement multi-view images and micro-lens image arrays for LFSD (Fig. \ref{Generated micro-lens image and multi-view images of LFSD}).
This complete data makes more comprehensive model evaluation possible. For example, models based on focal stacks, such as MoLF and ERNet, can now run and test on Lytro Illum. Note that for the remaining DUTLF-FS and DUTLF-MV, supplementing more data is possible in the future if the authors release the raw/more data. If this has been done, DUTLF-FS/DUTLF-MV has the potential to be the \emph{standard training dataset} for future models thanks to its large scale.

\begin{table*}[t!]
    \caption{Quantitative measures: S-measure ($S_\alpha$)~\cite{Fan2017StructureMeasureAN}, max F-measure ($F_{\beta}^{\textrm{max}}$), mean F-measure ($F_{\beta}^{\textrm{mean}}$)~\cite{Achanta2009FrequencytunedSR}, adaptive F-measure ($F_{\beta}^{\textrm{adp}}$)~\cite{Achanta2009FrequencytunedSR}, max E-measure ($E_{\phi}^{\textrm{max}}$), mean E-measure ($E_{\phi}^{\textrm{mean}}$)~\cite{Fan2018EnhancedalignmentMF}, adaptive E-measure ($E_{\phi}^{\textrm{adp}}$)~\cite{Achanta2009FrequencytunedSR} and MAE ($M$)~\cite{Perazzi2012SaliencyFC} of nine light field SOD models (\ie, LFS~\cite{Li2014SaliencyDOLFS}, WSC~\cite{Li2015WSC}, DILF~\cite{Zhang2015SaliencyDILF}, RDFD~\cite{Wang2020RegionbasedDRDFD}, DLSD~\cite{Piao2019DeepLSDLSD}, MoLF~\cite{Zhang2019MemoryorientedDFMoLF}, ERNet~\cite{Piao2020ExploitARERNet}, LFNet~\cite{Zhang2020LFNetLFNet}, MAC~\cite{Zhang2020LightFSLFDCN}) and nine SOTA RGB-D based SOD models (\ie, BBS~\cite{Fan2020BBSNetRS}, JLDCF~\cite{Fu2020JLDCFJL,Fu2021siamese}, SSF~\cite{Zhang2020SelectSA}, UCNet~\cite{Zhang2020UCNetUI}, D3Net~\cite{Fan2020RethinkingRS}, S2MA~\cite{LiuLearningSS}, cmMS~\cite{Li2020RGBDSO}, HDFNet~\cite{Pang2020HierarchicalDF}, and ATSA~\cite{Zhang2020ASTA}). Note in the table, light field SOD models are marked by ``$^\dag$''. Symbol ``N/T'' indicates that a model was not tested. The top three models among light field and RGB-D based SOD models are highlighted in \textcolor{red}{red}, \textcolor{blue}{blue} and \textcolor{green}{green}, \textbf{separately}. $\uparrow$/$\downarrow$ denotes that a larger/smaller value is better.}
    \label{tab_sota}
   % \vspace{-0.3cm}
   \centering
    % \footnotesize
    \renewcommand{\arraystretch}{1.0}
    \setlength{\tabcolsep}{0.5mm}
    \resizebox{0.98\textwidth}{!}{
    \begin{tabular}{p{0.9mm}r||c|c|c|c|c|c|c|c|c|c|c|c|c|c|c|c|c|c}
    \hline
    %&   & \multicolumn{7}{c|}{Light Field SOD Models}
    %&\multicolumn{9}{c}{RGB-D SOD Models} \\
    \cline{3-20}
     &   & \multicolumn{4}{c|}{Traditional}
    &\multicolumn{14}{c}{Deep learning-based}\\
    %\multicolumn{9}{c}{Deep learning-based}  \\
    \hline

            & Metric  & \tabincell{c}{LFS$^\dag$\\\cite{Li2014SaliencyDOLFS}} & \tabincell{c}{WSC$^\dag$\\\cite{Li2015WSC}} & \tabincell{c}{DILF$^\dag$\\\cite{Zhang2015SaliencyDILF}}  & \tabincell{c}{RDFD$^\dag$\\\cite{Wang2020RegionbasedDRDFD}}  & \tabincell{c}{DLSD$^\dag$\\\cite{Piao2019DeepLSDLSD}} & \tabincell{c}{MoLF$^\dag$\\\cite{Zhang2019MemoryorientedDFMoLF}}  & \tabincell{c}{ERNet$^\dag$\\\cite{Piao2020ExploitARERNet}} & \tabincell{c}{LFNet$^\dag$\\\cite{Zhang2020LFNetLFNet}} & \tabincell{c}{MAC$^\dag$\\\cite{Zhang2020LightFSLFDCN}} & \tabincell{c}{BBS\\\cite{Fan2020BBSNetRS}} & \tabincell{c}{JLDCF\\\cite{Fu2020JLDCFJL}} & \tabincell{c}{SSF\\\cite{Zhang2020SelectSA}} & \tabincell{c}{UCNet\\\cite{Zhang2020UCNetUI}}& \tabincell{c}{D3Net\\\cite{Fan2020RethinkingRS}}& \tabincell{c}{S2MA\\\cite{LiuLearningSS}}&\tabincell{c}{cmMS\\\cite{Li2020RGBDSO}} &\tabincell{c}{HDFNet\\\cite{Pang2020HierarchicalDF}}& \tabincell{c}{ATSA\\\cite{Zhang2020ASTA}}\\
    \hline
    \hline
        \multirow{8}{*}{\begin{sideways}\textit{LFSD}\cite{Li2014SaliencyDOLFS}\end{sideways}}  & $S_\alpha\uparrow$       &0.681&0.702& 0.811 & 0.786 & 0.786&\textcolor{blue}{0.825}&\textcolor{red}{0.831}  & \textcolor{green}{0.820} & 0.789&\textcolor{red}{0.864}&\textcolor{blue}{0.862}&\textcolor{green}{0.859}&0.858&0.825&0.837&0.850 & 0.846 & 0.858\\
                                                                        & $F_{\beta}^{\textrm{max}}\uparrow$       &0.744&0.743&\textcolor{green}{0.811}& 0.802 & 0.784&\textcolor{blue}{0.824}&\textcolor{red}{0.842}  & \textcolor{blue}{0.824} & 0.788&0.858&\textcolor{blue}{0.867}&\textcolor{red}{0.868}&0.859&0.812&0.835&0.858 & 0.837 & \textcolor{green}{0.866}\\
                                                                        & $F_{\beta}^{\textrm{mean}}\uparrow$
                                                                        & 0.513 & 0.722 & 0.719 & 0.735 & 0.758 & \textcolor{blue}{0.800} & \textcolor{red}{0.829}   & \textcolor{green}{0.794} &  0.753 & 0.842 & 0.848 & \textcolor{red}{0.862} & 0.848 & 0.797 & 0.806 & \textcolor{green}{0.850} & 0.818 & \textcolor{blue}{0.856} \\
                                                                        & $F_{\beta}^{\textrm{adp}}\uparrow$     &0.735&0.743& 0.795 & 0.802 &0.779&\textcolor{blue}{0.810}&\textcolor{red}{0.831}& \textcolor{green}{0.806} &0.793&0.840&0.827&\textcolor{red}{0.862}  & 0.838&0.788&0.803&\textcolor{blue}{0.857}&0.818&\textcolor{green}{0.852}  \\
                                                                        & $E_{\phi}^{\textrm{max}}\uparrow$        &0.809&0.789& 0.861 & 0.851 & 0.859 & \textcolor{green}{0.880} &\textcolor{blue}{0.884}  & \textcolor{red}{0.885} & 0.836&\textcolor{green}{0.900}&\textcolor{red}{0.902}&\textcolor{blue}{0.901}&0.898&0.863&0.873&0.896 & 0.880 & \textcolor{red}{0.902}\\
                                                                        & $E_{\phi}^{\textrm{mean}}\uparrow$       &0.567&0.753&0.764& 0.758 & 0.819 &\textcolor{green}{0.864}&\textcolor{red}{0.879}  & \textcolor{blue}{0.867} & 0.790&0.883&\textcolor{blue}{0.894}&0.890&\textcolor{green}{0.893}&0.850&0.855&0.881&0.869&\textcolor{red}{0.899}  \\
                                                                        & $E_{\phi}^{\textrm{adp}}\uparrow$      &0.773&0.788&0.846& 0.834 &\textcolor{green}{0.852}&\textcolor{blue}{0.879}&\textcolor{red}{0.882}  & \textcolor{red}{0.882} & 0.839&0.889&0.882&\textcolor{blue}{0.896}&\textcolor{green}{0.890}&0.853&0.863&\textcolor{green}{0.890}&0.872&\textcolor{red}{0.897}  \\
                                                                        & $M\downarrow$                            &0.205&0.150&0.136& 0.136 &\textcolor{green}{0.117}&\textcolor{blue}{0.092}&\textcolor{red}{0.083}  & \textcolor{blue}{0.092} & 0.118&0.072&\textcolor{green}{0.070}&\textcolor{red}{0.067}&0.072&0.095&0.094&0.073 & 0.086 & \textcolor{blue}{0.068}\\
    \hline
        \multirow{8}{*}{\begin{sideways}\textit{HFUT-Lytro}\cite{zhang2017MA}\end{sideways}}                   & $S_\alpha\uparrow$        &0.565&0.613&0.672& 0.619 &0.711&\textcolor{blue}{0.742}&\textcolor{red}{0.778}  & \textcolor{green}{0.736} & 0.731 &0.751&\textcolor{red}{0.789}&0.725&0.748&0.749&0.729&0.723 & \textcolor{green}{0.763} & \textcolor{blue}{0.772}\\
                                                                        & $F_{\beta}^{\textrm{max}}\uparrow$       &0.427&0.508&0.601 & 0.533 & 0.624&\textcolor{green}{0.662}&\textcolor{red}{0.722}  & 0.657 & \textcolor{blue}{0.667}&0.676&\textcolor{blue}{0.727}&0.647&0.677&0.671&0.650&0.626 & \textcolor{green}{0.690} & \textcolor{red}{0.729}\\
                                                                        & $F_{\beta}^{\textrm{mean}}\uparrow$
                                                                        & 0.323 & 0.493 & 0.513  & 0.469 & 0.594 & \textcolor{blue}{0.639} & \textcolor{red}{0.709}   & \textcolor{green}{0.628} &  0.620 & 0.654 & \textcolor{red}{0.707} & 0.639 & 0.672 & 0.651 & 0.623 & 0.617 & \textcolor{green}{0.669} & \textcolor{blue}{0.706} \\
                                                                        & $F_{\beta}^{\textrm{adp}}\uparrow$     &0.427&0.485&0.530 & 0.518 &0.592&\textcolor{green}{0.627}&\textcolor{red}{0.706}  & 0.615 & \textcolor{blue}{0.638}&0.654&\textcolor{blue}{0.677}&0.636&\textcolor{green}{0.675}&0.647&0.588&0.636&0.653&\textcolor{red}{0.689}  \\
                                                                        & $E_{\phi}^{\textrm{max}}\uparrow$        &0.637&0.695&0.748 & 0.712 &0.784&\textcolor{blue}{0.812}&\textcolor{red}{0.841}  & \textcolor{green}{0.799} & 0.797 &0.801&\textcolor{red}{0.844}&0.778&\textcolor{green}{0.804}&0.797&0.777&0.784 & 0.801 & \textcolor{blue}{0.833}\\
                                                                        & $E_{\phi}^{\textrm{mean}}\uparrow$       &0.524&0.684&0.657 & 0.623 & 0.749 &\textcolor{blue}{0.790}&\textcolor{red}{0.832}  & \textcolor{green}{0.777} & 0.733&0.765&\textcolor{red}{0.825}&0.763&\textcolor{green}{0.793}&0.773&0.756&0.746&0.788&\textcolor{blue}{0.819}  \\
                                                                        & $E_{\phi}^{\textrm{adp}}\uparrow$      &0.666&0.680&0.693 & 0.691 &0.755&\textcolor{blue}{0.785}&\textcolor{red}{0.831}  & 0.770 & \textcolor{green}{0.772}&\textcolor{green}{0.804}&\textcolor{red}{0.811}&0.781&\textcolor{blue}{0.810}&0.789&0.744&0.779&0.789&\textcolor{blue}{0.810}  \\
                                                                        & $M\downarrow$                            &0.221&0.154&0.150 & 0.223 &0.111&\textcolor{blue}{0.094}&\textcolor{red}{0.082}  & \textcolor{green}{0.103} &  0.107 &\textcolor{green}{0.089}&\textcolor{red}{0.075}&0.100&0.090&0.091&0.112&0.097 & 0.095 & \textcolor{blue}{0.084}\\

    \hline
    \multirow{8}{*}{\begin{sideways}\textit{Lytro Illum}\cite{Zhang2020LightFSLFDCN}\end{sideways}}                   & $S_\alpha\uparrow$       & 0.619 & 0.709 & 0.756  & 0.738 & \textcolor{green}{0.788} & \textcolor{blue}{0.834} & \textcolor{red}{0.843}   & N/T &  N/T & 0.879 & \textcolor{red}{0.890} & 0.872 & 0.865 & 0.869 & 0.853 & \textcolor{green}{0.881} & 0.873 & \textcolor{blue}{0.883} \\
 																		& $F_{\beta}^{\textrm{max}}\uparrow$	&	0.545 & 0.662 & 0.697  & 0.696 & \textcolor{green}{0.746} & \textcolor{blue}{0.820} & \textcolor{red}{0.827}   & N/T &  N/T & 0.850 & \textcolor{red}{0.878} & 0.850 & 0.843 & 0.843 & 0.823 & \textcolor{green}{0.857} & 0.855 & \textcolor{blue}{0.875}  \\
																		& $F_{\beta}^{\textrm{mean}}\uparrow$	&	0.385 & 0.646 & 0.604  & 0.624 & \textcolor{green}{0.713} & \textcolor{blue}{0.766} & \textcolor{red}{0.800}   & N/T &  N/T & 0.829 & \textcolor{red}{0.848} & \textcolor{green}{0.836} & 0.827 & 0.818 & 0.788 & \textcolor{blue}{0.839} & 0.823 & \textcolor{red}{0.848}  \\
	   																	& $F_{\beta}^{\textrm{adapt}}\uparrow$		& 0.547 & 0.639 & 0.659  & 0.679 & \textcolor{green}{0.720} & \textcolor{blue}{0.747} & \textcolor{red}{0.796}   & N/T &  N/T & 0.828 & 0.830 & \textcolor{blue}{0.835} & \textcolor{green}{0.824} & 0.813 & 0.778 & \textcolor{blue}{0.835} & 0.823 & \textcolor{red}{0.842}  \\
																		& $E_{\phi}^{\textrm{max}}\uparrow$		    & 0.721 & 0.804 & 0.830  & 0.816 & \textcolor{green}{0.871} & \textcolor{blue}{0.908} & \textcolor{red}{0.911}   & N/T &  N/T & 0.913 & \textcolor{red}{0.931} & 0.913 & 0.910 & 0.909 & 0.895 & \textcolor{green}{0.914} & 0.913 & \textcolor{blue}{0.929}  \\
																		& $E_{\phi}^{\textrm{mean}}\uparrow$ 		& 0.546 & 0.791 & 0.726  & 0.738 & \textcolor{green}{0.830} & \textcolor{blue}{0.882} & \textcolor{red}{0.900}   & N/T &  N/T & 0.900 & \textcolor{red}{0.919} & \textcolor{blue}{0.907} & \textcolor{green}{0.904} & 0.894 & 0.873 & \textcolor{blue}{0.907} & 0.898 & \textcolor{red}{0.919}  \\
																		& $E_{\phi}^{\textrm{adapt}}\uparrow$		& 0.771 & 0.797 & 0.812  & 0.815 & \textcolor{green}{0.855} & \textcolor{blue}{0.876} & \textcolor{red}{0.900}   & N/T &  N/T & 0.912 & 0.914 & \textcolor{red}{0.917} & \textcolor{green}{0.907} & \textcolor{green}{0.907} & 0.878 & \textcolor{blue}{0.915} & 0.904 & \textcolor{red}{0.917}  \\
																		& $M\downarrow$                  			& 0.197 & 0.115 & 0.132  & 0.142 & \textcolor{green}{0.086} & \textcolor{blue}{0.065} & \textcolor{red}{0.056}   & N/T &  N/T & 0.047 & \textcolor{blue}{0.042} & \textcolor{green}{0.044} & 0.048 & 0.050 & 0.063 & 0.045 & 0.051 & \textcolor{red}{0.041} \\

    \hline
        \multirow{8}{*}{\begin{sideways}\textit{DUTLF-FS } \cite{Wang2019DeepLFDLLF}\end{sideways}}                   & $S_\alpha\uparrow$        & 0.585 & 0.656 & 0.725  & 0.658 & N/T & \textcolor{blue}{0.887} & \textcolor{red}{0.899}   & \textcolor{green}{0.878} &  0.804 & 0.894 & \textcolor{green}{0.905} & \textcolor{red}{0.908} & 0.870 & 0.852 & 0.845 & \textcolor{blue}{0.906} & 0.868 & \textcolor{green}{0.905}  \\
 																		& $F_{\beta}^{\textrm{max}}\uparrow$		& 0.533 & 0.617 & 0.671  & 0.599 & N/T & \textcolor{blue}{0.903} & \textcolor{red}{0.908}   & \textcolor{green}{0.891} &  0.792 & 0.884 & \textcolor{blue}{0.908} & \textcolor{red}{0.915} & 0.864 & 0.840 & 0.829 & \textcolor{green}{0.906} & 0.857 & \textcolor{red}{0.915}  \\
																		& $F_{\beta}^{\textrm{mean}}\uparrow$		& 0.358 & 0.607 & 0.582  & 0.538 & N/T & \textcolor{blue}{0.855} & \textcolor{red}{0.891}   & \textcolor{green}{0.843} &  0.746 & 0.867 & 0.885 & \textcolor{red}{0.907} & 0.854 & 0.820 & 0.806 & \textcolor{green}{0.893} & 0.841 & \textcolor{blue}{0.899}  \\
	   																	& $F_{\beta}^{\textrm{adapt}}\uparrow$		& 0.525 & 0.617 & 0.663  & 0.599 & N/T & \textcolor{blue}{0.843} & \textcolor{red}{0.885}   & \textcolor{green}{0.831} &  0.790 & 0.872 & 0.874 & \textcolor{red}{0.903} & 0.850 & 0.826 & 0.791 & \textcolor{green}{0.887} & 0.835 & \textcolor{blue}{0.893}  \\
																		& $E_{\phi}^{\textrm{max}}\uparrow$		    & 0.711 & 0.788 & 0.802  & 0.774 & N/T & \textcolor{blue}{0.939} & \textcolor{red}{0.949}   & \textcolor{green}{0.930} &  0.863 & 0.923 & \textcolor{blue}{0.943} & \textcolor{red}{0.946} & 0.909 & 0.891 & 0.883 & \textcolor{green}{0.936} & 0.898 & \textcolor{blue}{0.943}  \\
																		& $E_{\phi}^{\textrm{mean}}\uparrow$ 		& 0.511 & 0.759 & 0.695  & 0.686 & N/T & \textcolor{blue}{0.921} & \textcolor{red}{0.943}   & \textcolor{green}{0.912} &  0.806 & 0.908 & \textcolor{green}{0.932} & \textcolor{red}{0.939} & 0.904 & 0.874 & 0.866 & 0.928 & 0.889 & \textcolor{blue}{0.938}  \\
																		& $E_{\phi}^{\textrm{adapt}}\uparrow$		& 0.742 & 0.787 & 0.813  & 0.782 & N/T & \textcolor{blue}{0.923} & \textcolor{red}{0.943}   & \textcolor{green}{0.913} &  0.872 & 0.924 & 0.930 & \textcolor{red}{0.942} & 0.905 & 0.895 & 0.870 & \textcolor{green}{0.931} & 0.895 & \textcolor{blue}{0.936} \\
																		& $M\downarrow$                  			& 0.227 & 0.151 & 0.156  & 0.191 & N/T & \textcolor{blue}{0.051} & \textcolor{red}{0.039}   & \textcolor{green}{0.054} &  0.102 & 0.054 & 0.043 & \textcolor{red}{0.036} & 0.059 & 0.069 & 0.079 & \textcolor{green}{0.041} & 0.065 & \textcolor{blue}{0.039}  \\

\hline
\end{tabular}}
%\vspace{-0.4cm}
\end{table*}

\begin{table*}[t]
	\centering
	\caption{Quantitative measures: S-measure ($S_\alpha$)~\cite{Fan2017StructureMeasureAN}, max F-measure ($F_{\beta}^{\textrm{max}}$), max E-measure ($E_{\phi}^{\textrm{max}}$), and MAE ($M$)~\cite{Perazzi2012SaliencyFC} of one retrained light field SOD model (ERNet~\cite{Piao2020ExploitARERNet}) and seven retrained RGB-D based SOD models (\ie, BBS~\cite{Fan2020BBSNetRS}, SSF~\cite{Zhang2020SelectSA}, ATSA~\cite{Zhang2020ASTA},  S2MA~\cite{LiuLearningSS}, D3Net~\cite{Fan2020RethinkingRS}, HDFNet~\cite{Pang2020HierarchicalDF}, and JLDCF~\cite{Fu2020JLDCFJL,Fu2021siamese}). Note in the table, the results of original models are taken from Table \ref{tab_sota}, and the retrained models are marked by ``*''. The best results of retrained models are highlighted in \textbf{bold}. $\uparrow$/$\downarrow$ denotes that a larger/smaller value is better.}
	\label{tab:quantitative}
% 	\vspace{2pt}
	\renewcommand{\arraystretch}{1.2}
	\setlength\tabcolsep{2pt}
	\resizebox{0.98\textwidth}{!}{
		\begin{tabular}{l||cccc||cccc||cccc||cccc}
			\hline \hline
% 			\toprule
             \multirow{2}{*}{{Models}} &
			  \multicolumn{4}{c||}{\tabincell{c}{LFSD~\cite{Li2014SaliencyDOLFS}}}
			& \multicolumn{4}{c||}{ \tabincell{c}{HFUT-Lytro~\cite{zhang2017MA}}}
			& \multicolumn{4}{c||}{\tabincell{c}{Lytro Illum~\cite{Zhang2020LightFSLFDCN}}}
			& \multicolumn{4}{c}{\tabincell{c}{DUTLF-FS~\cite{Wang2019DeepLFDLLF}}}\\
			\cline{2-17}
			& $S_\alpha\uparrow$ & $F_{\beta}^{\textrm{max}}\uparrow$ & $E_{\phi}^{\textrm{max}}\uparrow$ & $M\downarrow$
			& $S_\alpha\uparrow$ & $F_{\beta}^{\textrm{max}}\uparrow$ & $E_{\phi}^{\textrm{max}}\uparrow$ & $M\downarrow$
			& $S_\alpha\uparrow$ & $F_{\beta}^{\textrm{max}}\uparrow$ & $E_{\phi}^{\textrm{max}}\uparrow$ & $M\downarrow$
			& $S_\alpha\uparrow$ & $F_{\beta}^{\textrm{max}}\uparrow$ & $E_{\phi}^{\textrm{max}}\uparrow$ & $M\downarrow$
			\\
			\hline
% 			\\ \hline
% 			\input{table_data.tex}
	
	    BBS~\cite{Fan2020BBSNetRS}      & 0.864 & 0.858 & 0.900 & 0.072 & 0.751 & 0.676 & 0.801 & 0.089 & 0.879 & 0.850 & 0.913 & 0.047 & 0.894 & 0.884 & 0.923 & 0.054 \\
        SSF~\cite{Zhang2020SelectSA}      & 0.859 & 0.868 & 0.901 & 0.067 & 0.725 & 0.647 & 0.778 & 0.100 & 0.872 & 0.850 & 0.913 & 0.044 & 0.908 & 0.915 & 0.946 & 0.036 \\
        ATSA~\cite{Zhang2020ASTA}     & 0.858 & 0.866 & 0.902 & 0.068 & 0.772 & 0.729 & 0.833 & 0.084 & 0.883 & 0.875 & 0.929 & 0.041 & 0.905 & 0.915 & 0.943 & 0.039 \\
        ERNet~\cite{Piao2020ExploitARERNet}    & 0.831 & 0.842 & 0.884 & 0.083 & 0.778 & 0.722 & 0.841 & 0.082 & 0.843 & 0.827 & 0.911 & 0.056 & 0.899 & 0.908 & 0.949 & 0.039 \\
        S2MA~\cite{LiuLearningSS}     & 0.837 & 0.835 & 0.873 & 0.094 & 0.729 & 0.650 & 0.777 & 0.112 & 0.853 & 0.823 & 0.895 & 0.063 & 0.845 & 0.829 & 0.883 & 0.079 \\
        D3Net~\cite{Fan2020RethinkingRS}    & 0.825 & 0.812 & 0.863 & 0.095 & 0.749 & 0.671 & 0.797 & 0.091 & 0.869 & 0.843 & 0.909 & 0.050 & 0.852 & 0.840 & 0.891 & 0.069 \\
        HDFNet~\cite{Pang2020HierarchicalDF}   & 0.846 & 0.837 & 0.879 & 0.086 & 0.763 & 0.690 & 0.801 & 0.095 & 0.873 & 0.855 & 0.913 & 0.051 & 0.868 & 0.857 & 0.898 & 0.065 \\
        JLDCF~\cite{Fu2020JLDCFJL}    & 0.862 & 0.867 & 0.902 & 0.070 & 0.789 & 0.727 & 0.844 & 0.075 & 0.890 & 0.878 & 0.931 & 0.042 & 0.905 & 0.908 & 0.943 & 0.043 \\

        \hline

        BBS*~\cite{Fan2020BBSNetRS}     & 0.739 & 0.738 & 0.812 & 0.123 & 0.708 & 0.622 & 0.773 & 0.113 & 0.825 & 0.788 & 0.878 & 0.065 & 0.873 & 0.870 & 0.919 & 0.051 \\

        SSF*~\cite{Zhang2020SelectSA}     & 0.790 & 0.793 & 0.861 & 0.097 & 0.687 & 0.612 & 0.781 & 0.112 & 0.833 & 0.799 & 0.886 & 0.059 & 0.881 & 0.889 & 0.930 & 0.050 \\

        ATSA*~\cite{Zhang2020ASTA}    & 0.816 & 0.823 & 0.873 & 0.087 & 0.727 & 0.673 & 0.805 & 0.104 & 0.844 & 0.822 & 0.905 & 0.054 & 0.880 & 0.892 & 0.936 & 0.045 \\

        ERNet*~\cite{Piao2020ExploitARERNet}   & 0.822 & 0.825 & 0.885 & 0.085 & 0.707 & 0.632 & 0.766 & 0.128 & 0.840 & 0.810 & 0.900 & 0.059 & 0.898 & 0.903 & 0.946 & 0.040 \\

        S2MA*~\cite{LiuLearningSS}    & 0.827 & 0.829 & 0.873 & 0.086 & 0.672 & 0.572 & 0.735 & 0.133 & 0.839 & 0.802 & 0.885 & 0.060 & 0.894 & 0.893 & 0.934 & 0.046 \\

        D3Net*~\cite{Fan2020RethinkingRS}   & 0.827 & 0.821 & 0.877 & 0.086 & 0.720 & 0.645 & 0.801 & 0.104 & 0.859 & 0.835 & 0.906 & 0.051 & 0.906 & 0.911 & 0.947 & 0.039 \\

        HDFNet*~\cite{Pang2020HierarchicalDF}  & 0.849 & 0.850 & 0.891 & 0.073 & 0.747 & 0.673 & 0.801 & \textbf{0.096} & 0.874 & 0.854 & 0.915 & 0.045 & 0.922 & \textbf{0.931} & 0.955 & \textbf{0.030} \\

        JLDCF*~\cite{Fu2020JLDCFJL}   & \textbf{0.850} & \textbf{0.860} & \textbf{0.900} & \textbf{0.071} & \textbf{0.755} & \textbf{0.694} & \textbf{0.823} & 0.098 & \textbf{0.877} & \textbf{0.855} & \textbf{0.919} & \textbf{0.042} & \textbf{0.924} & \textbf{0.931} & \textbf{0.958} & \textbf{0.030} \\

    \hline
		\end{tabular}
	}
\end{table*}

\subsection{Performance Benchmarking and Analysis}
\label{Performance Benchmarking and Analysis}

\begin{figure*}[!ht]
\centering
\includegraphics[width=0.98\textwidth]{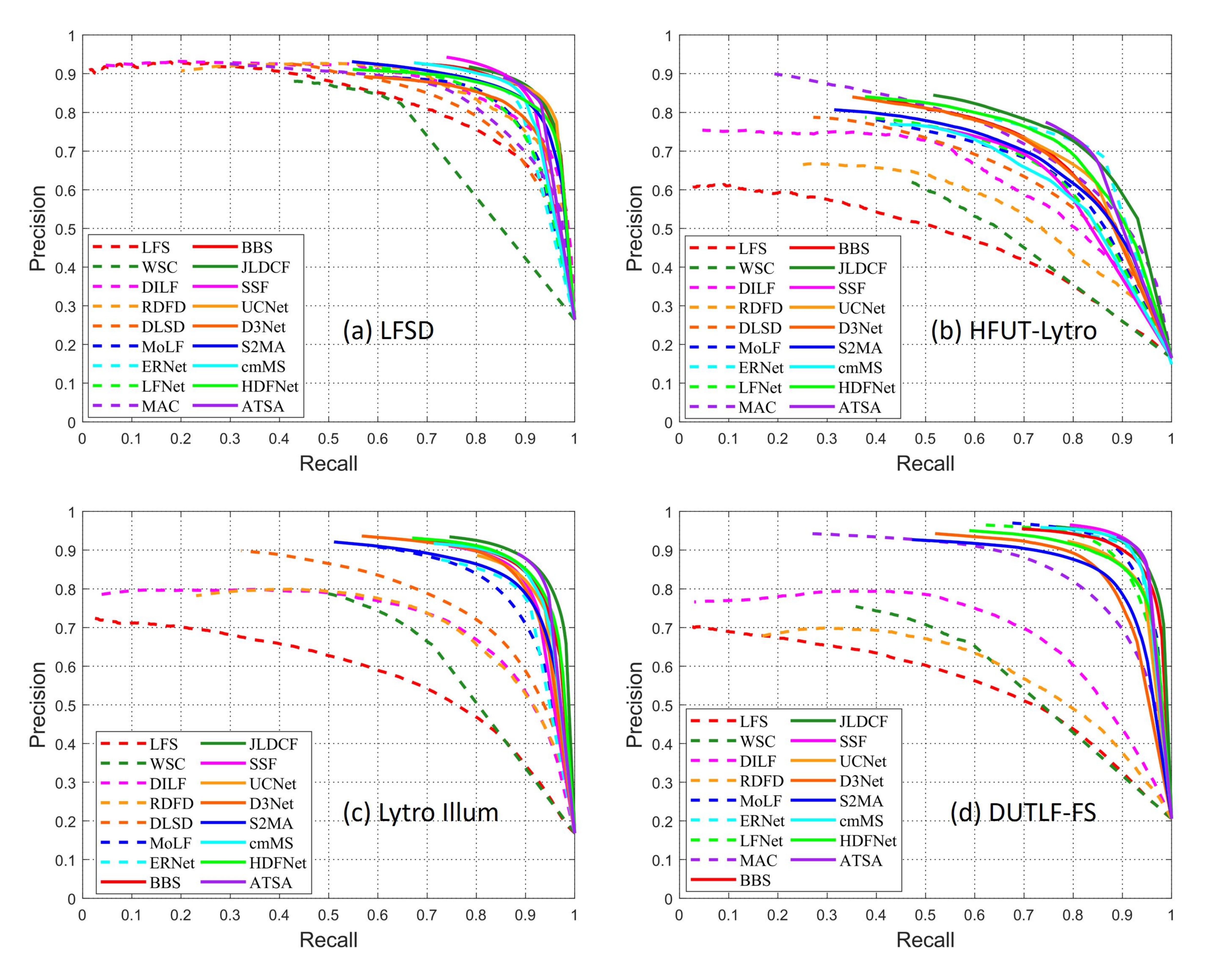}
\caption{PR curves on four datasets ((a) LFSD~\cite{Li2014SaliencyDOLFS}, (b) HFUT-Lytro~\cite{zhang2017MA}, (c) Lytro Illum~\cite{Zhang2020LightFSLFDCN}, and (d) DUTLF-FS~\cite{Wang2019DeepLFDLLF}) for nine light field SOD models (\ie, LFS~\cite{Li2014SaliencyDOLFS}, WSC~\cite{Li2015WSC}, DILF~\cite{Zhang2015SaliencyDILF}, RDFD~\cite{Wang2020RegionbasedDRDFD}, DLSD~\cite{Piao2019DeepLSDLSD}, MoLF~\cite{Zhang2019MemoryorientedDFMoLF}, ERNet~\cite{Piao2020ExploitARERNet}, LFNet~\cite{Zhang2020LFNetLFNet}, MAC~\cite{Zhang2020LightFSLFDCN}) and nine SOTA RGB-D based SOD models (\ie, BBS~\cite{Fan2020BBSNetRS}, JLDCF~\cite{Fu2020JLDCFJL,Fu2021siamese}, SSF~\cite{Zhang2020SelectSA}, UCNet~\cite{Zhang2020UCNetUI}, D3Net~\cite{Fan2020RethinkingRS}, S2MA~\cite{LiuLearningSS}, cmMS~\cite{Li2020RGBDSO}, HDFNet~\cite{Pang2020HierarchicalDF}, and ATSA~\cite{Zhang2020ASTA}). Note that in this figure, the \emph{solid lines} and \emph{dashed lines} represent the PR curves of \emph{RGB-D based SOD models} and \emph{light field SOD models}, respectively.}
\label{PR curve}
\end{figure*}

\begin{figure*}[!ht]
\centering
\includegraphics[width=0.98\textwidth]{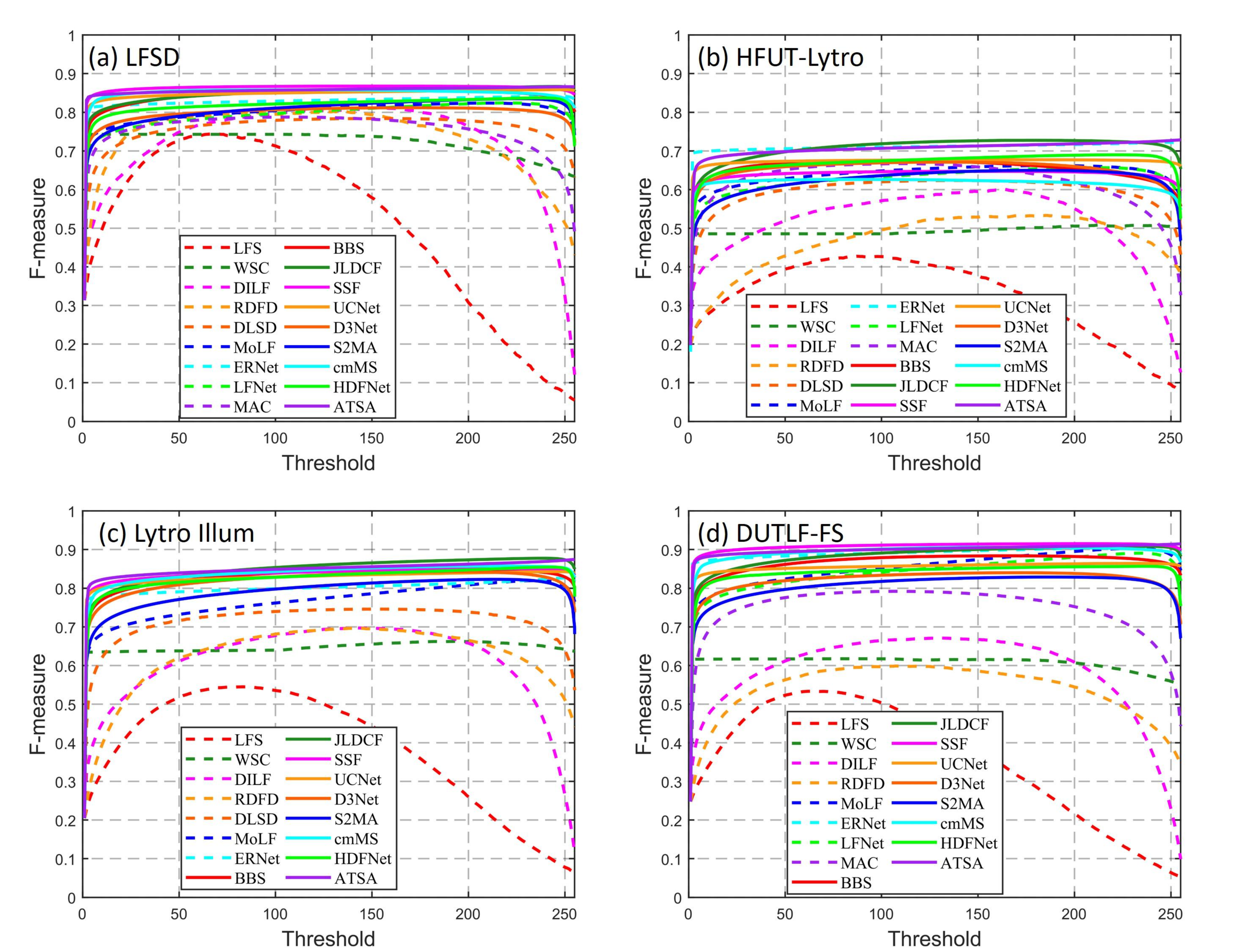}
\caption{F-measure curves under different thresholds on four datasets ((a) LFSD~\cite{Li2014SaliencyDOLFS}, (b) HFUT-Lytro~\cite{zhang2017MA}, (c) Lytro Illum~\cite{Zhang2020LightFSLFDCN} and (d) DUTLF-FS~\cite{Wang2019DeepLFDLLF}) for nine light field SOD models (\ie, LFS~\cite{Li2014SaliencyDOLFS}, WSC~\cite{Li2015WSC}, DILF~\cite{Zhang2015SaliencyDILF}, RDFD~\cite{Wang2020RegionbasedDRDFD}, DLSD~\cite{Piao2019DeepLSDLSD}, MoLF~\cite{Zhang2019MemoryorientedDFMoLF}, ERNet~\cite{Piao2020ExploitARERNet}, LFNet~\cite{Zhang2020LFNetLFNet}, MAC~\cite{Zhang2020LightFSLFDCN}) and nine SOTA RGB-D based SOD models (\ie, BBS~\cite{Fan2020BBSNetRS}, JLDCF~\cite{Fu2020JLDCFJL,Fu2021siamese}, SSF~\cite{Zhang2020SelectSA}, UCNet~\cite{Zhang2020UCNetUI}, D3Net~\cite{Fan2020RethinkingRS}, S2MA~\cite{LiuLearningSS}, cmMS~\cite{Li2020RGBDSO}, HDFNet~\cite{Pang2020HierarchicalDF}, and ATSA~\cite{Zhang2020ASTA}). Note that in this figure, the \emph{solid lines} and \emph{dashed lines} represent the F-measure curves of \emph{RGB-D based SOD models} and \emph{light field SOD models}, respectively.}
\label{Fmeasure curve}
\end{figure*}

To provide an in-depth understanding of the performance of different models, we conduct the first comprehensive benchmarking of nine light field SOD models (\ie, LFS~\cite{Li2014SaliencyDOLFS}, WSC~\cite{Li2015WSC}, DILF~\cite{Zhang2015SaliencyDILF}, RDFD \cite{Wang2020RegionbasedDRDFD} DLSD~\cite{Piao2019DeepLSDLSD}, MoLF~\cite{Zhang2019MemoryorientedDFMoLF}, ERNet~\cite{Piao2020ExploitARERNet}, LFNet \cite{Zhang2020LFNetLFNet} MAC~\cite{Zhang2020LightFSLFDCN}) and nine SOTA RGB-D based SOD models\footnote{The RGB-D SOD models benchmarked in this paper are selected according to the top models concluded by the recent survey~\cite{Zhou2020RGBDSO} and also the latest open-source models published in ECCV-2020. All depth maps fed to a model are optionally reversed on an entire dataset to fit the best performance of this model.
% All RGB-D SOD models are not re-trained in our experiments, and instead just use their released model weights trained on RGB-D data. This generally evaluates how well they can generalize to the light field images.
} (\ie, BBS~\cite{Fan2020BBSNetRS}, JLDCF~\cite{Fu2020JLDCFJL,Fu2021siamese}, SSF~\cite{Zhang2020SelectSA}, UCNet~\cite{Zhang2020UCNetUI}, D3Net~\cite{Fan2020RethinkingRS}, S2MA~\cite{LiuLearningSS}, cmMS~\cite{Li2020RGBDSO}, HDFNet~\cite{Pang2020HierarchicalDF}, and ATSA~\cite{Zhang2020ASTA}) on four existing light field datasets, including the entire LFSD (100 light fields), HFUT-Lytro (255 light fields), Lytro Illum (640 light fields) datasets and the test set (462 light fields) of DUTLF-FS. Sample images from these datasets are shown in Fig. \ref{rgb, depth and GT}. All the benchmarked models have either publicly available source/executable codes or results provided by the authors (the authors of RDFD \cite{Wang2020RegionbasedDRDFD} and LFNet \cite{Zhang2020LFNetLFNet} have sent us the saliency map results).
The nine evaluation metrics described previously (\ie, PR, S-measure, max/mean F-measure, max/mean E-measure, adaptive F-measure and E-measure, mean absolute error) are adopted and the results are reported in Table \ref{tab_sota}. Meanwhile, the PR curves, max F-measure curves and visual comparisons are shown in Fig. \ref{PR curve}, Fig. \ref{Fmeasure curve}, Fig. \ref{visual comparison} and Fig. \ref{visual comparison on small and multiple objects}, respectively.

It is worth noting that the evaluation is not conducted on the DUTLF-MV dataset~\cite{Piao2019DeepLSDLSD} since it only provides multi-view images, which are not compatible with the input data forms of most light field SOD models.
Further, note that DLSD~\cite{Piao2019DeepLSDLSD} is not tested on the DUTLF-FS test set because it used quite some test images ($\sim$36\%, according to our verification) from this dataset for training.
Also, MAC \cite{Zhang2020LightFSLFDCN} is not evaluated on Lytro Illum since the authors conducted five-fold cross-validation on this dataset, and therefore it is not directly comparable to other models.
Besides, for DUTLF-FS has no available micro-lens image arrays after dataset unification as illustrated in Table \ref{completion table}, and the quality of micro-lens image arrays of HFUT-Lytro is fairly low due to the low-quality multi-view images, we follow~\cite{Zhang2020LightFSLFDCN} and instead test MAC on single up-sampled all-in-focus images from these two datasets.  %###########
%As for LFSD, we evaluated MAC on the generated micro-lens image arrays of it, and the saliency maps are finally warp to the ground-truth of all-in-focus images for fair evaluation.  %###########
In addition, for ERNet~\cite{Piao2020ExploitARERNet}, we only evaluate the teacher model since its pre-trained student model is not publicly available.
%Also, as the ERNet used 100 samples of HFUT-Lytro for training, we only evaluated it on the remaining 155 samples of HFUT-Lytro. %###########
Comprehensive analyses are given as follows.

\subsubsection{Traditional \emph{vs.} Deep Light Field SOD Models}
Compared to the four traditional models shown in Table \ref{model table}, the deep learning-based SOD models clearly have a significant performance boost on all datasets. The best traditional model evaluated, namely DILF, is generally inferior to any deep model. This confirms the power of deep neural networks when applied to this field.

\subsubsection{Deep Learning-Based Light Field SOD Models} \
\label{Comparison on deep models}
As shown in Table \ref{model table}, MoLF, ERNet and LFNet adopt focal stacks and all-in-focus images as input data forms, while DLSD and MAC handle center-view images and micro-lens image arrays. From Table \ref{tab_sota} and Fig. \ref{PR curve}, it is clear that MoLF, ERNet and LFNet outperform DLSD and MAC. It is also worth noting that MoLF and ERNet achieve the top-2 performance, which is probably because that these two models were trained on the large-scale DUTLF-FS dataset (\ie, 1,000 light fields) with superior network structures. Besides, such results indicate that models based on multi-view or micro-lens images are not as effective as those based on focal stacks. This is probably because that the former are less studied, and the effectiveness of multi-view and micro-lens images is still underexplored. Moreover, the training data may also matter because MAC was trained only on Lytro Illum, which is about half the scale of DUTLF-FS. Among the above five models compared, ERNet attains the top-1 accuracy.

\subsubsection{Comparison Between Light Field SOD and RGB-D SOD Models} From the quantitative results illustrated in Table \ref{tab_sota} and Fig. \ref{PR curve}, it can be observed that, the latest cutting-edge RGB-D models achieve comparable or even better performance than the light field SOD models. Notably, three RGB-D models, namely JLDCF, SSF, and ATSA, achieve generally better performance than ERNet on most datasets. The underlying reasons may be two-fold. First, RGB-D based SOD has recently drawn extensive research interest and many powerful and elaborate models have been proposed. Inspired by previous research on the RGB SOD problem~\cite{Wu2019StackedCR,Feng2019AttentiveFN,Qin2019BASNetBS}, these models often pursue edge-preserving results from deep neural networks and employ functional modules/architectures, such as boundary supplement unit~\cite{Zhang2020SelectSA}, multi-scale feature aggregation module~\cite{Zhang2020ASTA}, or UNet-shaped bottom-up/top-down architecture~\cite{Fu2020JLDCFJL,LiuLearningSS,Li2020ICNetIC}. In contrast, light field SOD has been less explored and the models/architectures evolve slowly. Edge-aware properties have not yet been considered by most existing models. For example, although the attention mechanism and ConvLSTM are adopted in ERNet, no UNet-like top-down refinement is implemented to generate edge-aware saliency maps. As evidenced in Fig. \ref{light field and RGB-D examples} and Fig. \ref{visual comparison}, the RGB-D SOD models tend to detect more accurate boundaries than existing deep light field SOD models. Second, the other potential reason could be that the RGB-D SOD models are trained on more data. For instance, the universally agreed training set for the RGB-D SOD task contains 2,200 RGB-D scenarios \cite{Fu2020JLDCFJL}, while ERNet~\cite{Piao2020ExploitARERNet} was trained only on $\sim$1,000 light fields. Thus, the former is more likely to obtain better generalizability.

However, we can still hardly deny the potentials of light fields on boosting the performance of SOD, as recently RGB-D SOD area is much more active (many new competitive models are proposed as mentioned in~\cite{Zhou2020RGBDSO}) than the area of light field SOD. Besides, the performance of ERNet and MoLF is only slightly lower than that of the RGB-D models on the benchmark datasets, which further implies the effectiveness of light fields for SOD~\cite{Zhang2015LightFSCS}. We believe that there is still considerable room for improving light field SOD, because light fields can provide more information than paired RGB and depth images.

Furthermore, in order to eliminate training discrepancy, we conduct experiments by retraining those RGB-D models on a unified training set, namely the training set of DUTLF-FS that contains 1,000 scenarios. We also retrain ERNet to remove its extra HFUT-Lytro training data as shown in Table \ref{model table}. Comparative results are given in Table \ref{tab:quantitative}, where all the models generally incur certain performance degeneration. Interestingly, after retraining, SSF* can no longer outperform ERNet*, while ATSA* becomes inferior to ERNet* on LFSD and DUTLF-FS. Only JLDCF* and HDFNet* consistently outperform ERNet* with a noticeable margin.

\subsubsection{Accuracy Across Different Datasets}
\label{Comparison on different datasets}
It is clearly shown in Table \ref{tab_sota} and Fig. \ref{PR curve} that the models tested perform differently on different datasets. Generally, the models achieve better results on LFSD than on the other three datasets, indicating that LFSD is the easiest dataset for light field SOD, on which the traditional model DILF can even outperform some deep models like DLSD and MAC. In contrast, HFUT-Lytro, Lytro Illum and DUTLF-FS are more challenging.
%From Fig. \ref{rgb, depth and GT}, it is also clear that, with the exception of LFSD, the datasets have limited light field depth ranges.
Note that MoLF, ERNet, ATSA behave prominently on DUTLF-FS, probably because they were trained on DUTLF-FS's training set or training data (Table \ref{model table}).
Besides, as mentioned in Section \ref{Light field SOD datasets}, HFUT-Lytro has many small salient objects, with multiple objects existed per image. The degraded performance of these models on this dataset tells that detecting small/multiple salient objects is still very challenging for existing schemes, no matter for RGB-D based models or light field models. This makes HFUT-Lytro the most difficult among existing light field datasets.

\subsubsection{Results Visualization}
\label{Results visualization}
\begin{figure*}[!htbp]
    \centering
    \includegraphics[width=0.95\textwidth]{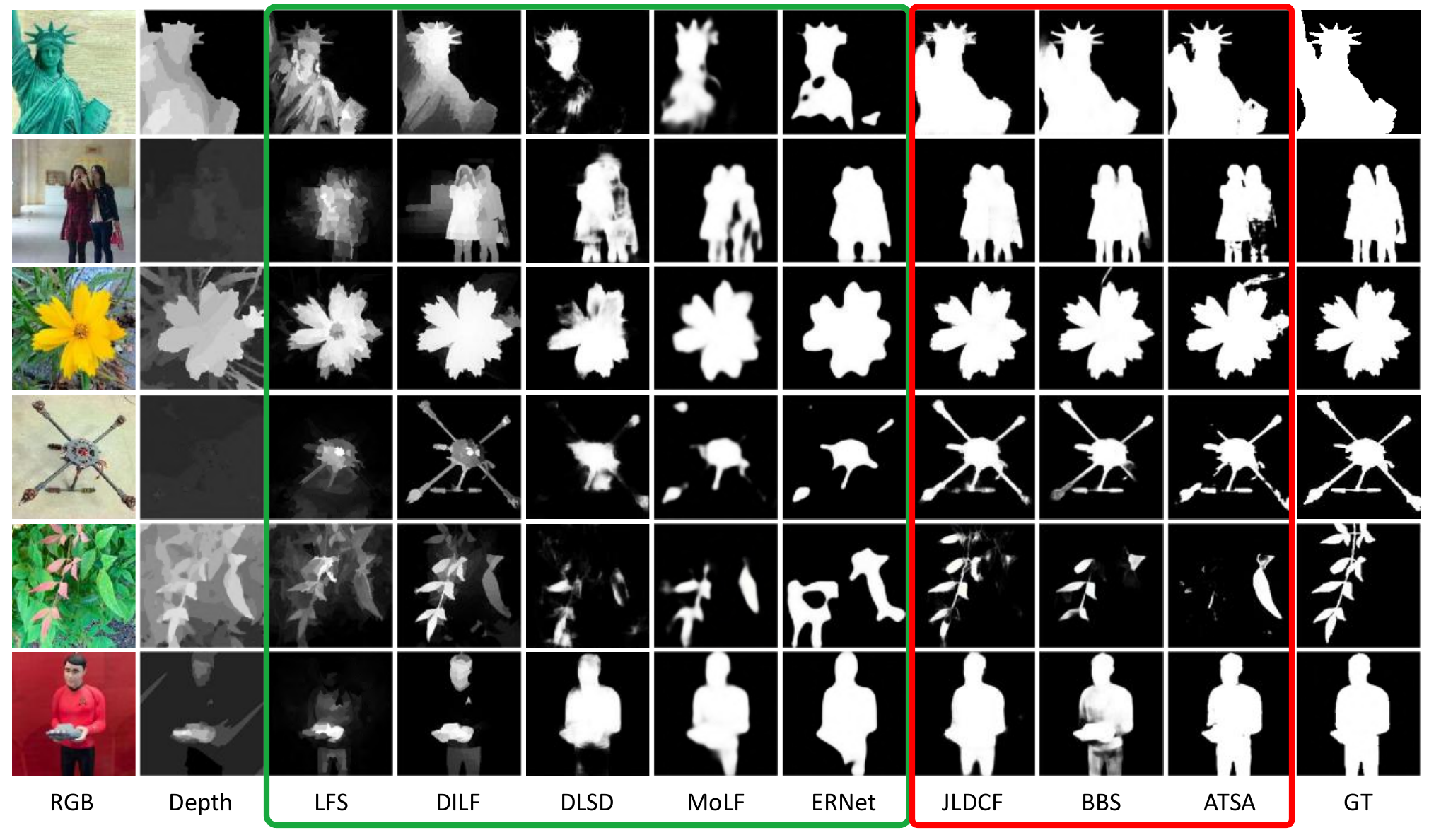}%\vspace{-0.3cm}
    \caption{Visual comparison of five light field SOD (\ie, LFS~\cite{Li2014SaliencyDOLFS}, DILF~\cite{Zhang2015SaliencyDILF}, DLSD~\cite{Piao2019DeepLSDLSD}, MoLF~\cite{Zhang2019MemoryorientedDFMoLF} and ERNet~\cite{Piao2020ExploitARERNet}, bounded in the green box) and three SOTA RGB-D based SOD models (\ie, JLDCF~\cite{Fu2020JLDCFJL,Fu2021siamese}, BBS~\cite{Fan2020BBSNetRS}, and ATSA~\cite{Zhang2020ASTA}, bounded in the red box).}
    \label{visual comparison}
\end{figure*}

\begin{figure*}[!ht]
    \centering
    \includegraphics[width=0.95\textwidth]{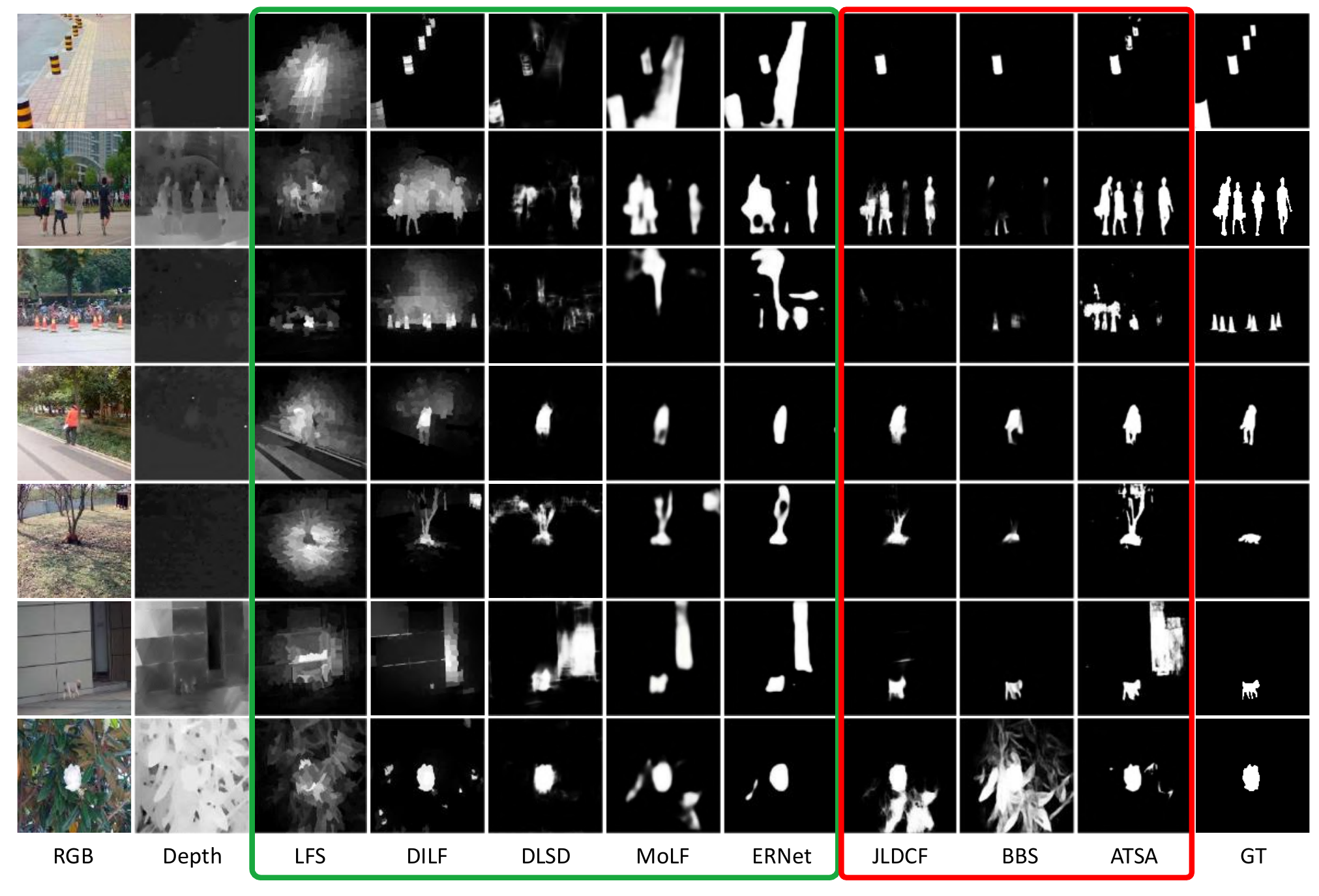}\vspace{-0.3cm}
    \caption{Visual comparison of five light field SOD (\ie, LFS~\cite{Li2014SaliencyDOLFS}, DILF~\cite{Zhang2015SaliencyDILF}, DLSD~\cite{Piao2019DeepLSDLSD}, MoLF~\cite{Zhang2019MemoryorientedDFMoLF} and ERNet~\cite{Piao2020ExploitARERNet}, bounded in the green box.) and three SOTA RGB-D based SOD models (\ie, JLDCF~\cite{Fu2020JLDCFJL,Fu2021siamese}, BBS~\cite{Fan2020BBSNetRS}, and ATSA~\cite{Zhang2020ASTA}, bounded in the red box) on detecting small and multiple objects.}
    \label{visual comparison on small and multiple objects}
\end{figure*}

Fig. \ref{visual comparison} visualizes some sample results from five light field models, including two traditional methods (\ie, LFS and DILF) and three deep learning-based models (\ie, DLSD, MoLF and ERNet ), and three latest RGB-D based models (\ie, JLDCF, BBS, and ATSA). The first two rows in Fig. \ref{visual comparison} show easy cases while the third to fifth rows show cases with complex backgrounds or sophisticated boundaries. The last row gives an example with low color contrast between foreground and background.
As can be observed from Fig. \ref{visual comparison}, RGB-D models
perform comparably to or even better than light field models, which confirms the fact that this field is still insufficiently studied.
Fig. \ref{visual comparison on small and multiple objects} further shows several scenarios with small and multiple salient objects, where the first three rows show the cases with multiple salient objects and the others show the cases of small objects. According to Fig. \ref{visual comparison on small and multiple objects}, both RGB-D based and light field models are more likely to result in erroneous detection, confirming the challenge of handling small/multiple objects for existing techniques.

\section{Challenges and Open Directions}
\label{challenge and open direction}
This section highlights several future research directions for light field SOD and outlines several open issues.

\subsection{Dataset Collection and Unification}
As demonstrated in Section~\ref{Light field SOD datasets}, existing light field datasets are limited in scale and have non-unified data forms, making it somewhat difficult to evaluate different models and generalize deep networks. This non-unified issue is \emph{particularly severe} for light field SOD because of its diverse data representations and high dependency on special acquisition hardware, differing it from other SOD tasks (\eg, RGB-D SOD~\cite{Fu2020JLDCFJL,Zhang2020SelectSA,Zhang2020ASTA}, video SOD~\cite{Tsiami2020STAViSSA,Fan2019ShiftingMA}) in the saliency community. Therefore, developing large-scale and unified datasets is essential for future research. We urge researchers to take this issue into consideration when constructing new datasets. Moreover, collecting complete data forms, including raw data, focal stacks, multi-view images, depth maps, and micro-lens image arrays, would definitely facilitate and advance research on this topic. However, we also note there is a challenge in data storage and transmission, since raw light field data is quite large in size (\eg, the 640 light fields of Lytro Illum occupy 32.8 Gigabytes), not to mention a large-scale dataset.
The scale of the dataset makes it a bit difficult to spread. In this case, it will still be great if a subset of any data form is available for the public.

\subsection{Further Investigations in Light Field SOD}
As discussed, there are currently fewer studies regarding SOD on light fields compared to other tasks in the saliency community. Thus, this field is still less active and under-explored. Besides, from the benchmarking results in Section \ref{Performance Benchmarking and Analysis}, it can be observed that the SOTA performance is still far from satisfactory, especially on the HFUT-Lytro dataset. There is considerable room for further improvement of light field SOD algorithm/model. In addition, we can notice that only seven deep learning-based models appeared between 2019 and 2020. We attribute such a scarcity of light field SOD research to the aforementioned data problems, as well as the lack of a complete survey of existing methods and datasets on this topic.

%Considerable room exists for further improvement on algorithms/models. Further, we can note that only seven deep learning-based models emerged between 2019 and 2020. We attribute such scarce research on light field SOD to the data issue mentioned previously, as well as the lack of a complete survey of existing methods and datasets on this topic, which is rightly the goal of this paper.

\subsection{Multi-view Images and Micro-lens Image Arrays}

Most existing models work with focal stacks and depth maps, as summarized in Table \ref{model table}, while multi-view images and micro-lens image arrays are two other types of light field data representations that are rarely considered (only five related models). The benchmarking results in Section \ref{Performance Benchmarking and Analysis} show that the related models do not perform as well as models utilizing other data forms, so the use of these two data forms has not yet been fully explored. Thus, it is expected to develop more light field SOD models in the future for exploring the effectiveness of multi-view images and micro-lens image arrays. Another potential reason could be that, these two data forms themselves may be less information-representative than focal stacks and depth maps, namely scene depth information is more implicitly conveyed. This may make it difficult to find effective mappings and mine underlying rules using deep neural networks, especially when the training data is sparse. In addition, it is an interesting work to compare the effectiveness and redundancy of saliency detection using different data forms.

\subsection{Incorporating High-Quality Depth Estimation}
It has been proven that accurate depth maps are conducive to discovering salient objects from complex backgrounds. Unfortunately, the quality of depth maps varies greatly in several existing datasets, since depth estimation from light fields is a challenging task~\cite{Jeon2015AccurateDM,Tao2013DepthFC,Tao2015DepthFS,Wang2015OcclusionAwareDE,Piao2021DynamicFN,Piao2021LearningMI}. %easily leading to imperfect depth.
The challenge stems from the fact that although the light fields can be used to synthesize images focused at any depth through digital refocusing technology, the depth distribution of each scene point is unknown. Besides, it is necessary to determine whether the image area is in focus, which itself is an unresolved issue~\cite{Zhao2020DefocusBD,Park2017AUA}. Imperfect depth maps often result in negative effects on the detection accuracy of models using depth maps. Therefore, incorporating high-quality depth estimation algorithms from light field is for sure beneficial for boosting performance.

\subsection{Edge-Aware Light Field SOD}
Accurate object boundaries are essential for high-quality saliency maps, as SOD is a pixel-wise segmentation task~\cite{Borji2015SalientOD}. In the RGB SOD field, edge-aware SOD
models are drawing increasing research attention~\cite{Wu2019StackedCR,Feng2019AttentiveFN,Qin2019BASNetBS}. Currently, as shown in our experimental results, existing deep light field SOD models rarely consider this issue, resulting in saliency maps with coarse boundaries/edges. Thus, edge-aware light field SOD could be a future research direction.

\subsection{Development of Acquisition Technology and Hardware}
As mentioned before, the first generation light field camera, named Lytro, was invented in 2011, while its successor, called Lytro Illum, was introduced in 2014. The latter is more powerful but has a much large size than the former, and is also much more expensive. However, in general, the development of light field acquisition technology and hardware has been slower than that of IT fields such as computers and mobile phones. Since 2014, there are few commercial light field cameras. There is an urgent need for the development of acquisition and hardware technology for light field photography.
Currently, light field cameras are far from replacing traditional RGB cameras in terms of image quality, price and portability.
Imagining that in the future if the light field cameras become affordable and small in size, which can easily be integrated into people's mobile phones. In this case, everyone can try light field photography in daily life. This would provide a vast increase in user data and post-processing application requirements, paving the way for significant improvements in light field SOD.

\subsection{Different Supervision Strategies}
Existing deep light field models learn to segment salient objects in a fully supervised manner, which requires sufficient annotated training data. Unfortunately, as mentioned before, the size of the existing datasets is limited, \emph{e.g.}, DUTLF-FS and DUTLF-MV provide 1,000 and 1,100 samples for training, respectively, while other datasets contain fewer than 640 light fields. On one hand, a small amount of training data limits the generalization ability of models. On the other hand, acquiring a large amount of annotated data requires extreme manual effort on data collection and labelling. Recently, weakly- and semi-supervised learning strategies have attracted extensive research attention, largely reducing the annotation effort. Because of being data-friendly, they have been introduced to RGB SOD and some encouraging attempts~\cite{Zeng2019MultiSourceWS,Zhang2016BridgingSD,Qian2019LanguageawareWS}
have been made. Inspired by this, one future direction is to extend these supervision strategies to light field SOD, which can overcome the shortage of training data. Additionally, several works \cite{Chen2020AdversarialRF,Dai2020SGNNSG} have proven that pre-training models in a self-supervised manner can effectively improve performance, which can also be introduced to light field SOD in the future.

\subsection{Linking RGB-D SOD to Light Field SOD}
There is a close connection between light field SOD and RGB-D SOD, since both tasks explore scene depth information for saliency detection, whereas depth information can be derived from light field data via different techniques. This is why RGB-D SOD can be regarded as a solution to the degradation of light field SOD. As shown in Table \ref{tab_sota}, applying RGB-D SOD models to light field is straightforward, whereas we believe its reverse could also be possible. For example, intuitively, reconstructing light field data such as focal stacks or multi-view images from a pair of RGB and depth images is possible (\eg, \cite{Piao2019DeepLSDLSD}). If this bridge is realized, mutual transfer between the models of these two fields becomes feasible, and then light field models can be applied to RGB-D data.
Such a link would be an interesting issue to explore in the near future.
%\fkr{Such a link allows not only model transfer, but also data transfer between RGB-D SOD and light field SOD}.

\subsection{Other Potential Directions}
%There are several other potential directions for future research inspired by the recent advances in the saliency community.
Inspired by the recent advances in the saliency community, there are several other potential directions for future research.
For example, high-resolution salient object detection~\cite{Zeng2019TowardsHS} aims to deal with salient object segmentation in high-resolution images, thus achieving high-resolution details could be considered in light field SOD. Besides, while existing light field datasets are labelled at an object-level, instance-level annotation and detection, which aim at separating individual objects~\cite{Cai2019CascadeRH,Chen2019HybridTC,Liu2018PathAN,Li2017InstanceLevelSO,Fan2020S4NetSS}, could also be introduced into this field. Note that there are quite some instance-sensitive application scenarios, \emph{e.g.}, image captioning~\cite{Karpathy2017DeepVA}, and multi-label image recognition~\cite{Wei2014CNNST}, as well as various weakly supervised/unsupervised learning scenarios~\cite{Chen2015WeblySL,Lai2016SaliencyGD}. One recent work attempted to address weakly-supervised salient instance detection~\cite{Tian2020WeaklysupervisedSI}. Similarly, more effort could be spent on instance-level ground-truth annotation and designing instance-level light field SOD models. Furthermore, as we know, eye-fixation prediction~\cite{Borji2019SaliencyPI,Borji2015SalientOD,Borji2013StateoftheArtIV} is another subfield of saliency detection. So far, there has been no research on eye-fixation prediction using light field data. As abundant natural scene information is provided by the light field, we hope that the various data forms of the light field, which provide useful cues to help eliminate ambiguous eye-fixation. Lastly, light field data could benefit other closely related tasks to SOD, such as camouflaged object detection (COD) \cite{Fan2020CamouflagedOD} and transparent object segmentation \cite{Xu2015TransCutTO}, where objects often borrow texture from their background and have similar appearances to their surroundings.

Finally, there is an unanswered question remaining: \emph{how can light field information benefit SOD over depth information}? As we all know, depth information can be derived from and is a subset of light field data. Different forms of light field data, \emph{e.g.}, focal stacks and multi-view images, somewhat imply depth information, indicating that existing models may leverage such depth information in an implicit way. So how is the gap between using depth in an explicit way (like RGB-D SOD models do) and in an implicit way? This is an interesting question, but unfortunately, since the problem of light field SOD was proposed in 2014, no study has shown any direct answer/evidence. Such an answer is worthy of further investigation and understanding into this field in the future.

\section{Conclusions}\label{sec:conclusions}

We have provided the first comprehensive review and benchmark for light field SOD, reviewing and discussing existing studies and related datasets. We have benchmarked representative light field SOD models and compared them to several cutting-edge RGB-D SOD models both qualitatively and quantitatively. Considering the fact that the existing light field datasets are somewhat inconsistent in data forms, we generated supplemental data for existing datasets, making them complete and unified. Moreover, we have discussed several potential directions for future research and outlined some open issues. Although progress has been made over the past several years, there are still only seven deep learning-based works focusing on this topic, leaving significant room for designing more powerful network architectures and incorporating effective modules (\eg, edge-aware designs and top-down refinement) for improving SOD performance. We hope this survey will serve as a catalyst to advance this area and promote interesting works in the future.

%============================================%

% or


% \appendices

% \section{}

%\section*{Acknowledgment}

\ifCLASSOPTIONcaptionsoff
  \newpage
\fi



%\bibliographystyle{IEEEtran}
% \begin{thebibliography}[1]
\bibliographystyle{IEEEtran}
% % argument is your BibTeX string definitions and bibliography database(s)

%~\cite{Li2014SaliencyDO}
\bibliography{citation}
%~\cite{Li2014SaliencyDO}
%~\cite{Li2015AWS}
% \end{thebibliography}
% Li, N., Ye, J., Ji, Y., Ling, H., & Yu, J. (2014). Saliency Detection on Light Field. IEEE Transactions on Pattern Analysis and Machine Intelligence, 39, 1605-1616.
% \end{thebibliography}



\vfill


\end{document}